\documentclass{article} 
\usepackage{iclr2027_conference,times}

\iclrfinalcopy  
\renewcommand{\lhead}[1]{}
\renewcommand{\headrulewidth}{0pt}

\usepackage{amsmath,amsfonts,bm}

\def\eqref#1{equation~\ref{#1}}

\def\1{\bm{1}}

\DeclareMathAlphabet{\mathsfit}{\encodingdefault}{\sfdefault}{m}{sl}
\SetMathAlphabet{\mathsfit}{bold}{\encodingdefault}{\sfdefault}{bx}{n}

\usepackage{hyperref}
\usepackage{url}

\usepackage{booktabs}
\usepackage{graphicx}
\usepackage{amsmath}
\usepackage{wrapfig}
\usepackage{tipa}
\usepackage{makecell}
\usepackage{multirow}
\usepackage{pifont}
\usepackage[table]{xcolor}
\definecolor{lightgray}{gray}{0.93}
\usepackage{caption}
\usepackage{algorithm}
\usepackage{algpseudocode}
\usepackage{CJKutf8}
\usepackage{enumitem}

\title{Towards Multi-View Sign Language Understanding: A Benchmark Dataset and Baseline}

\author{%
  Xu Wang$\quad$Shengeng Tang\thanks{Corresponding author.}\hspace{0.45em}$\quad$Wan Jiang$\quad$Yaxiong Wang$\quad$Lechao Cheng$\quad$Richang Hong \\[0.35em]
  \normalfont\small Hefei University of Technology$\quad$
  \normalfont\small (wangxu2002@mail.hfut.edu.cn, tangsg@hfut.edu.cn)
}

\begin{document}

\maketitle

\begin{center}
  \vspace{-10mm}
  \includegraphics[width=\linewidth]{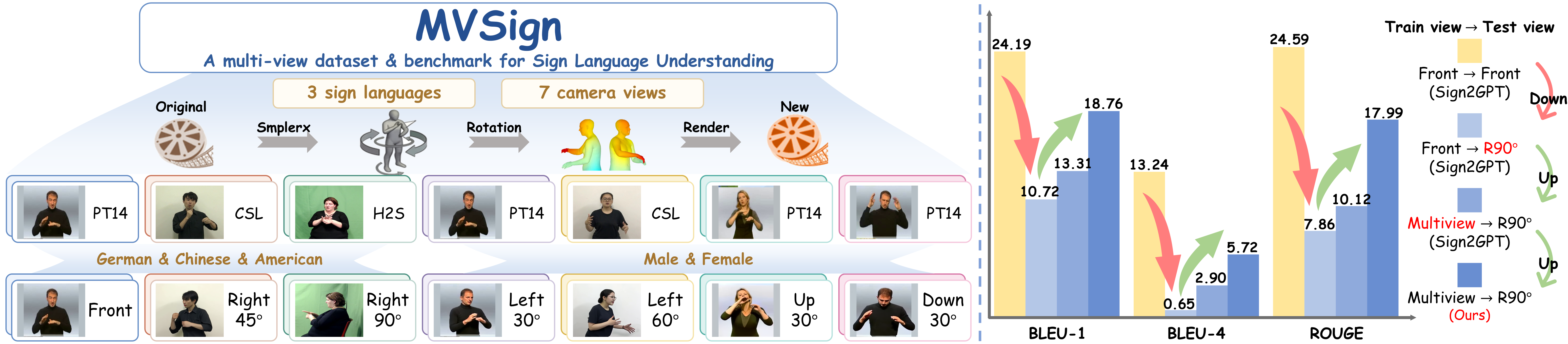}
  \vspace{-4mm}
  \captionof{figure}{
    Overview of MVSign benchmark and viewpoint robustness analysis.
    \textbf{Left:} MVSign combines original recordings with synthesized views to cover three sign languages and seven viewpoints.
    \textbf{Right:} A controlled comparison on How2Sign with a fixed training-set size. Sign2GPT degrades when testing shifts from frontal videos to paired native $R90^\circ$ recordings. Replacing 40\% of frontal training samples with multi-view samples improves side-view performance, while CanonSLU achieves further gains using the same multi-view training data and $R90^\circ$ test set.
  }
  \label{fig:overall}
\end{center}

\begin{abstract}
Most existing Sign Language Understanding (SLU) methods are developed under fixed-view settings and remain vulnerable to viewpoint changes, while the scarcity of multi-view data hinders systematic research on viewpoint robustness. To address this gap, we introduce \textbf{MVSign}, a benchmark spanning three sign languages and seven viewpoints. MVSign combines original frontal recordings from existing datasets with additional views synthesized via $3$D whole-body motion reconstruction and target-view rendering, while preserving the source splits and annotations. We further propose \textbf{CanonSLU}, a framework for multi-view SLU. This framework employs a canonical-view-guided strategy, using the frontal view as a semantic anchor. Specifically, CanonSLU incorporates \emph{Canonical-View Knowledge Transfer (CKT)} to transfer canonical knowledge from the frontal view to non-frontal views, and \emph{Motion Relation Aggregation (MRA)} to aggregate motion-related features across frames and strengthen temporal representations under viewpoint changes. Extensive experiments on MVSign demonstrate competitive performance with gains on non-frontal views, establishing CanonSLU as an effective baseline for multi-view SLU.
\end{abstract}

\section{Introduction}
Sign Language Understanding (SLU)~\citep{li2025Uni-Sign} converts sign language videos into gloss sequences or spoken-language sentences, serving as an important technique for improving communication accessibility between the deaf community and the broader information environment. 
SLU mainly encompasses Isolated Sign Language Recognition (ISLR)~\citep{pu2024siformer,wang2025exploiting}, Continuous Sign Language Recognition (CSLR)~\citep{hu2023continuous,gan2023contrastive,xu2025hierarchical} and Sign Language Translation (SLT)~\cite{wong2024sign2gpt,gan2025mixsigngraph,guo2025bridging}.
ISLR classifies individual sign movements, whereas CSLR learns the temporal alignment between continuous sign language sequences and their corresponding glosses. SLT further translates sign language videos into natural language sentences. Despite substantial progress on standard CSLR and SLT benchmarks~\citep{huang2018video,camgoz2018neural,zhou2021improving}, driven by advances in visual representation learning~\citep{koller2019weakly,min2021visual}, sequence modeling~\citep{hao2021self,gan2024signgraph,cheng2020fully}, and large-scale pre-training~\citep{chen2022simple,hu2023signbert+,li2025Uni-Sign}, most existing systems are developed and evaluated under nearly fixed-view settings, where training and testing videos are captured from the same or very similar camera positions. Consequently, their performance on current benchmarks may overestimate real-world robustness, leaving \textbf{\emph{viewpoint variation}} as a practical yet underexplored challenge.

We conduct a controlled experiment on How2Sign~\citep{duarte2021how2sign} with the representative SLT model Sign2GPT~\citep{wong2024sign2gpt}, using paired frontal and native $R90^\circ$ recordings (Figure~\ref{fig:overall} (right)). With the training-set size fixed, replacing the frontal-view test samples with their $R90^\circ$ side-view counterparts leads to a large performance drop, because viewpoint changes alter hand appearance, motion projection, occlusion, and hand-body coordination.
Replacing 40\% of the frontal training samples with multi-view samples improves side-view performance under the same training budget, showing that broader viewpoint coverage provides useful training signals for cross-view SLU. Moreover, CanonSLU achieves further gains on the same multi-view training data and $R90^\circ$  test set, suggesting that robust multi-view SLU requires not only viewpoint coverage but also models that exploit cross-view semantic and motion cues. However, such coverage is scarce in continuous SLU: mainstream benchmarks such as PHOENIX-2014T~\citep{camgoz2018neural} and CSL-Daily~\citep{zhou2021improving} mainly provide frontal-view videos; existing large-scale multi-view datasets~\citep{li2020word,desai2023asl,shen2024mm,dinh2025sign} primarily target ISLR; and How2Sign only provides frontal and $R90^\circ$ views without gloss annotations. Collecting synchronized multi-camera data with sequence-level annotations is also costly. These observations point to the need for multi-view data that can be obtained reliably and efficiently from existing single-view recordings, as well as models tailored to multi-view SLU.

To address this need, we introduce \textbf{MVSign}, a multi-view SLU benchmark spanning three sign languages and seven viewpoints. As illustrated in Figure~\ref{fig:overall} (left), MVSign comprises MVSign-PT14, MVSign-CSL, and MVSign-H2S. 
It uses the frontal view as a canonical reference shared by all sequences and extends it with six non-frontal views $\{R90^\circ, R45^\circ,L30^\circ,L60^\circ,U30^\circ,D30^\circ\}$. For sources without native multi-view recordings, new views are constructed around the signer, not the camera: each recording is first lifted into a unified $3$D whole-body motion representation, which then guides the re-rendering of new viewpoints into realistic videos, preserving the source splits, annotations, and motion content. 
To improve viewpoint robustness in multi-view SLU, we further propose \textbf{CanonSLU}, a canonical-view-guided framework that treats the frontal view as a semantic anchor. 
CanonSLU incorporates \emph{Canonical-View Knowledge Transfer (CKT)} to transfer canonical knowledge from a pretrained frontal-view model to non-frontal inputs, and \emph{Motion Relation Aggregation (MRA)} to aggregate motion-aware relations across frames and strengthen temporal representations under viewpoint changes. 
Our contributions are summarized as follows:

\begin{enumerate}[label=(\arabic*), leftmargin=2.8em, itemsep=0em, topsep=0em]
    \item We construct \textbf{MVSign}, a multi-view SLU benchmark covering three sign languages and seven viewpoints, combining synthesized and naturally captured multi-view videos while preserving original splits, annotations, and motion sequences.

    \item We propose \textbf{CanonSLU}, a canonical-view-guided framework that uses the frontal view as a semantic anchor, integrating CKT for frontal-to-non-frontal knowledge transfer and MRA for motion-aware temporal aggregation under viewpoint changes.
    
    \item Extensive experiments on MVSign across CSLR and SLT show that CanonSLU improves performance under viewpoint variations, with gains on non-frontal views, establishing a strong baseline for multi-view SLU.

\end{enumerate}

\section{MVSign Dataset}
\label{sec:mvsign}
\begin{figure*}[t]
  \centering
  \includegraphics[width=\linewidth]{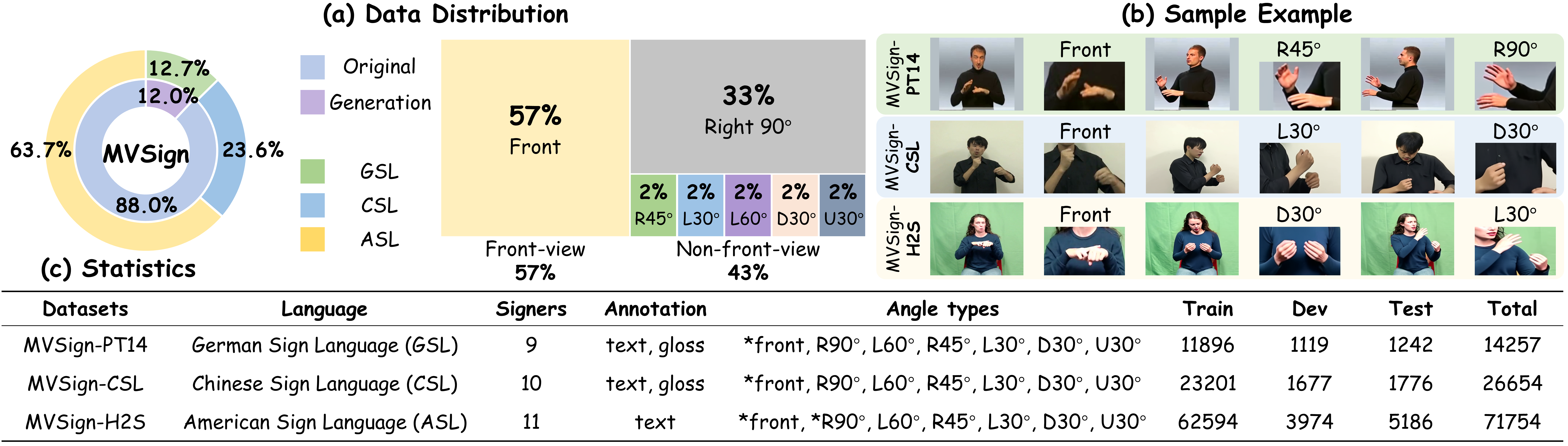}
  \vspace{-5mm}
  \caption{
  Overview of the MVSign dataset.
  (a) Data distribution over original vs. generated, sign languages and viewpoints.
  (b) Sample multi-view examples. 
  (c) Statistics of dataset.
  }
  \label{fig:data_distribution}
  \vspace{-3mm}
\end{figure*}

To expand viewpoint coverage for continuous SLU, we construct \textbf{MVSign}, a multi-view benchmark comprising MVSign-PT14, MVSign-CSL, and MVSign-H2S, derived from PHOENIX-2014T, CSL-Daily, and How2Sign, respectively. 
We retain all original frontal-view videos as the \emph{Front} view and preserve the native \emph{R90$^\circ$} recordings in MVSign-H2S.
Using the three-stage construction process in Figure~\ref{fig:main}, we generate 1,000 videos for each of six non-frontal viewpoints (\emph{R45$^\circ$}, \emph{R90$^\circ$}, \emph{L30$^\circ$}, \emph{L60$^\circ$}, \emph{U30$^\circ$}, and \emph{D30$^\circ$}) for both MVSign-PT14 and MVSign-CSL, yielding 6,000 additional videos per subset. For each of the five additional viewpoints in MVSign-H2S, excluding \emph{R90$^\circ$}, we sample 100 source videos from each original train, dev, and test split and apply the same construction process, yielding 300 generated videos per viewpoint. All generated videos inherit their source split assignments and linguistic annotations: gloss and sentence annotations for MVSign-PT14 and MVSign-CSL, and sentence annotations for MVSign-H2S. Figure~\ref{fig:data_distribution} summarizes the data distribution, examples, and statistics of MVSign.

\textbf{Whole-body Parameter Extraction.}
Given an input single-view RGB video, we first recover a 3D parameterized whole-body representation for each frame using SMPLer-X~\citep{cai2023smpler}, which is pretrained on multiple human behavior datasets (\emph{e.g.}, 3DPW~\citep{von2018recovering}, COCO~\citep{jin2020whole}, InterHand2.6M~\citep{moon2020interhand2} and UBody~\citep{lin2023one}). The estimated parameters are denoted by $\theta_s = \{\theta_b, \theta_h, \theta_e, \theta_j, \theta_{\mathrm{root}}, \theta_{\mathrm{shape}}, \psi\}$, 
where $\theta_b$, $\theta_h$, $\theta_e$, $\theta_j$, $\theta_{\mathrm{root}}$, $\theta_{\mathrm{shape}}$, $\psi$ represent the body pose, bilateral hand pose, eye pose, jaw pose, global root orientation, body shape, and facial expression parameters, respectively.

Based on these parameters, we reconstruct the intermediate mesh sequence. Compared with direct pixel-space manipulation, this parameterized 3D representation provides a more geometry-consistent basis for view synthesis and better preserves fine-grained sign-specific motion patterns under fast articulation, self-occlusion, and hand--body interaction.

\textbf{Root-Driven Viewpoint Transformation.}
After recovering the 3D body representation, we synthesize new viewpoints by transforming the body in 3D space rather than manipulating images directly.
We introduce viewpoint changes by modifying the global root ($\theta_{\mathrm{root}}$) orientation while retaining the estimated local pose, body-shape, and facial-expression parameters for each frame.
Let $\mathbf{V} \in \mathbb{R}^{3 \times M}$ denote the mesh vertices of a frame, where $M$ is the number of mesh vertices. Given a target viewpoint parameterized by yaw angle $\alpha$ and pitch angle $\beta$, the transformed mesh is defined as:
\begin{equation}
\mathbf{V}'
=
\mathbf{R}_{\mathrm{view}}\mathbf{V}
=
\mathbf{R}_{y}(\alpha)\mathbf{R}_{x}(\beta)\mathbf{V}
=
\setlength{\arraycolsep}{3pt}
\begin{bmatrix}
\cos\alpha&0&\sin\alpha\\
0&1&0\\
-\sin\alpha&0&\cos\alpha
\end{bmatrix}
\begin{bmatrix}
1&0&0\\
0&\cos\beta&-\sin\beta\\
0&\sin\beta&\cos\beta
\end{bmatrix}
\mathbf{V}.
\end{equation}
The same target-view rotation is applied across all frames, preserving the local joint configurations and relative temporal dynamics among the torso, arms, and hands of the recovered motion.

\textbf{Skeleton-Guided Adaptive Rendering.}
To obtain realistic sign language videos after viewpoint transformation, 
we first render the viewpoint-transformed mesh sequence to obtain an intermediate video. DWPose~\citep{yang2023effective} extracts a 2D skeleton sequence from the rendered frames, as motion guidance for StableAnimator~\citep{tu2025stableanimator}. Conditioned on this skeleton sequence and identity cues from the original video, StableAnimator generates the target-view sign language video.

\section{Method}
\subsection{Preliminaries}
Given a sign language video captured under viewpoint $v$, we denote the input sequence as $X_i^v=\{x_{i,t}^v\}_{t=1}^{T_i}\in\mathbb{R}^{T_i\times 3\times H\times W}$, where $i$ indexes the sign sample, $T_i$ is the number of frames, and $H$ and $W$ denote the frame height and width, respectively.
$v\in\mathcal{V}$ belongs to a predefined set of viewpoints.
We use $f\in\mathcal{V}$ to denote the frontal view, which serves as the canonical view in our framework. 
Depending on the available annotations, each sample may be associated with a gloss sequence $G_i=\{g_{i,m}\}_{m=1}^{M_i}$ and a spoken-language sentence $S_i=\{s_{i,n}\}_{n=1}^{N_i}$.
In this work, we study multi-view SLU under three task settings: CSLR, SLT (including Gloss-based SLT and Gloss-free SLT). 
(1) \textbf{CSLR} maps a sign video $X_i^v$ to its corresponding gloss sequence $G_i$.
(2) \textbf{Gloss-based SLT} translates $X_i^v$ into a spoken-language sentence $S_i$ using gloss annotations as supervision during training.
(3) \textbf{Gloss-free SLT} translates $X_i^v$ into $S_i$ without using manually annotated glosses.

\begin{figure*}[t]
  \centering
  \includegraphics[width=\textwidth]{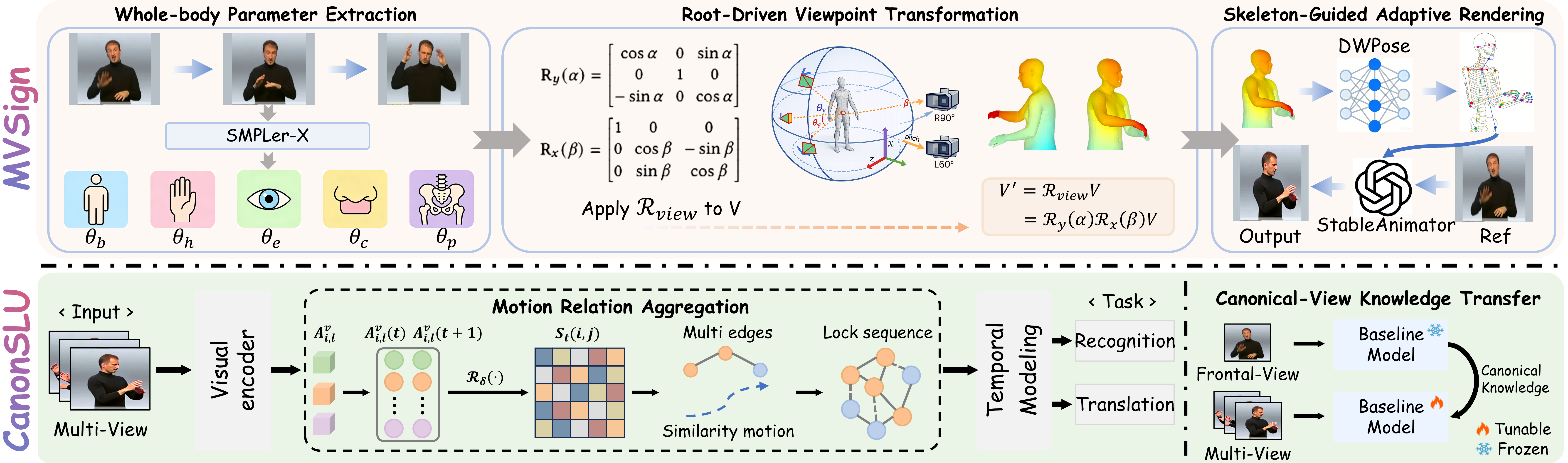}
  \vspace{-6mm}
  \caption{
  Overview of \textbf{MVSign} and \textbf{CanonSLU} framework.
  The upper row illustrates the three-stage construction process used to extend multi-view dataset.
  The lower row presents CanonSLU, which employs a shared visual-temporal encoder, \textbf{Motion Relation Aggregation}, and \textbf{Canonical-View Knowledge Transfer} with task-specific CSLR and SLT heads.
  }
  \label{fig:main}
\end{figure*}

\subsection{CanonSLU Framework}
Since viewpoint changes perturb hand appearance, motion projection, and occlusion patterns, they cause both representation-level disturbance and prediction-level semantic discrepancy. As shown in Figure~\ref{fig:main}, we build a framework for multi-view SLU, where a shared visual-temporal encoder supports both gloss and sentence prediction. 
CanonSLU consists of three components: (i) \textbf{Baseline Model} that extracts visual-temporal features and performs CSLR or SLT with task-specific prediction heads, (ii) Motion Relation Aggregation (\textbf{MRA}) for motion-aware representation learning within each view, (iii) Canonical-View Knowledge Transfer (\textbf{CKT}) that uses the frontal view as a semantic anchor and transfers canonical knowledge to non-frontal views when distillation is used.

\textbf{Baseline Model.}
Given a sign video $X_i^v$ captured under viewpoint $v$, we first extract frame-wise visual features with a visual encoder:
\begin{equation}
F_i^v = E_{\mathrm{vis}}(X_i^v).
\end{equation}
The visual features are then processed by a temporal convolution module and a BiLSTM, which capture short-range temporal dynamics and long-range temporal dependencies, respectively:
\begin{equation}
\widetilde{H}_i^v = \operatorname{CNN}_{1\mathrm{D}}(F_i^v), 
\qquad
H_i^v = \operatorname{BiLSTM}(\widetilde{H}_i^v).
\end{equation}
For \textbf{CSLR}, both the BiLSTM output $H_i^v$ and the convolution-level feature $\widetilde{H}_i^v$ are projected into the gloss vocabulary by the gloss classifier:
\begin{equation}
\widetilde{Z}_i^v = C_{\mathrm{gloss}}(\widetilde{H}_i^v),
\qquad
Z_i^v = C_{\mathrm{gloss}}(H_i^v).
\end{equation}
Here, $Z_i^v$ denotes the final sequence logits for gloss prediction, while $\widetilde{Z}_i^v$ provides auxiliary logits for intermediate supervision. The final gloss sequence is directly recognized from $Z_i^v$:
\begin{equation}
\hat{G}_i = \operatorname{Recognition}(Z_i^v).
\end{equation}
For \textbf{SLT}, the BiLSTM feature $H_i^v$ is first mapped into the mBART embedding space by a visual-language mapper, implemented as a two-layer MLP, which builds a connection between the two different modalities. The mapped feature is then fed into mBART to obtain translation logits:
\begin{equation}
T_i^v = \operatorname{mBART}\left(\operatorname{Mapper}(H_i^v)\right).
\end{equation}
Here, $T_i^v$ denotes the token-level logits over the text vocabulary. The final spoken sentence is directly translated from $T_i^v$:
\begin{equation}
\hat{S}_i = \operatorname{Translation}(T_i^v).
\end{equation}

\textbf{Motion Relation Aggregation.}
To improve viewpoint robustness at the representation level, we augment the visual encoder with \textbf{MRA}. 
The motivation is that viewpoint changes mainly affect local appearance and spatial projection, while the underlying action sequence remains unchanged.
Therefore, we model motion relations across frames and aggregate relevant visual features, aiming to capture more stable action cues across viewpoints for SLU.
Let $A_{i,l}^{v}\in\mathbb{R}^{C_l\times T_i\times H_l\times W_l}$ denote the intermediate feature map from the $l$-th stage of the visual encoder $E_{\mathrm{vis}}$. We first compute frame-wise motion features by temporal differencing:
\begin{equation}
M_{i,l}^v(t) = A_{i,l}^v(t) - A_{i,l}^v(t-1).
\end{equation}
Using the refined motion features and projected value features, MRA aggregates motion-related information across multiple temporal offsets:
\begin{equation}
O_{i,l}^{v}
=
\operatorname{Agg}_{\delta\in\mathcal{T}_{\mathrm{off}}}
\mathcal{R}_{\delta}
\left(
\phi_m(M_{i,l}^{v}),
\phi_v(A_{i,l}^{v})
\right),
\qquad
\mathcal{T}_{\mathrm{off}}=\{1,2,4\}.
\end{equation}
where $\phi_m(\cdot)$ is a temporal convolution for motion refinement, $\phi_v(\cdot)$ is a $1\times1\times1$ value projection, and $\mathcal{R}_{\delta}(\cdot)$ denotes motion relational aggregation over the temporal offsets $\delta$.
Specifically, it computes motion similarity between frames separated by each $\delta$, selects the most related spatial positions, and aggregates their value features in both forward and backward temporal directions.

Finally, the aggregated feature $O_{i,l}^{v}$ is selectively injected into the original feature map through a motion-aware gated residual update:
\begin{equation}
\widehat{A}_{i,l}^{v}
=
A_{i,l}^{v}
+
\alpha_l
\left[
\phi_o(O_{i,l}^{v}) \odot G_{i,l}^{v}
\right].
\end{equation}
Here, $\phi_o$ is an output temporal convolution, and $G_{i,l}^{v}$ is a sigmoid gate computed from the concatenation of $A_{i,l}^{v}$ and $\phi_m(M_{i,l}^{v})$.
The operator $\odot$ denotes element-wise multiplication, and $\alpha_l$ is a learnable residual weight.
In our implementation, MRA is applied at stages $l \in \{3,4\}$ of the visual encoder, where the lower spatial resolution reduces the cost of dense cross-frame matching.
The aggregation and gating operations are detailed in Appendix~\ref{sec:supp_mra}.

\textbf{Canonical-View Knowledge Transfer.}
Although MRA enhances motion-aware representations within each view, different viewpoints may still produce semantic biases. 
To reduce this discrepancy, we use the frontal view $f$ as the canonical semantic anchor and transfer its knowledge to non-frontal views.
Given paired frontal and non-frontal videos, we compute distillation at corresponding prediction positions and average the loss over valid positions:
\begin{equation}
\mathcal{D}_{\mathrm{KD}}(\mathcal{Q}_i^{f}, \mathcal{Q}_i^{v})
=
\mathbb{I}[v \neq f]
\frac{T_d^2}{\sum_{r=1}^{L_i}m_{i,r}}
\sum_{r=1}^{L_i}m_{i,r}\cdot
D_{\mathrm{KL}}
(
\sigma(\mathcal{Q}_{i,r}^{f}/T_d)
\parallel
\sigma(\mathcal{Q}_{i,r}^{v}/T_d)
),
\end{equation}
where $\mathcal{D}_{\mathrm{KD}}(\cdot,\cdot)$ denotes the logit-level distillation loss averaged over valid sequence positions.
$\mathcal{Q}_{i,r}^{f}$ and $\mathcal{Q}_{i,r}^{v}$ are the teacher and
student logits at position $r$, respectively.
$L_i$ denotes the sequence length including padding, and $m_{i,r}$ indicates a valid position.
The softmax $\sigma(\cdot)$ is applied over the task-specific
vocabulary, $T_d$ is the distillation temperature.
The position index $r$ refers to encoder time steps for CSLR and target-token positions for gloss-free SLT.
Teacher logits are detached from the computation graph, and $\mathbb{I}[v\neq f]$ activates distillation for non-frontal samples.

\subsection{Training Objective}
\textbf{CSLR.}
We first train a frontal-view recognition model using only frontal-view samples. The objective contains CTC losses on both the final sequence logits and the auxiliary convolutional logits:
\begin{equation}
\mathcal{L}_{\mathrm{CSLR}}^{\mathrm{front}}
=
\mathcal{L}_{\mathrm{CTC}}(Z_i^f,G_i)
+
\mathcal{L}_{\mathrm{CTC}}(\widetilde{Z}_i^f,G_i).
\end{equation}
Then, we train the model with all views and transfer frontal-view knowledge to non-frontal views:
\begin{equation}
\mathcal{L}_{\mathrm{CSLR}}^{\mathrm{multi}}
=
\mathcal{L}_{\mathrm{CTC}}(Z_i^v,G_i)
+
\mathcal{L}_{\mathrm{CTC}}(\widetilde{Z}_i^v,G_i)
+
\lambda
\mathcal{D}_{\mathrm{KD}}(Z_i^f,Z_i^v),
\end{equation}
where $Z_i^f$ denotes the pretrained sequence logits from the frontal-view recognition model.

\textbf{Gloss-based SLT.}
We initialize the visual-temporal encoder from the multi-view CSLR model trained with CKT and the translation module from mBART-large-cc25. Both modules are then jointly fine-tuned on all views using sentence-level cross-entropy and two gloss-level CTC losses:
\begin{equation}
\mathcal{L}_{\mathrm{GBSLT}}^{\mathrm{multi}}
=
\mathcal{L}_{\mathrm{CE}}(T_i^v,S_i)
+
\mathcal{L}_{\mathrm{CTC}}(Z_i^v,G_i)
+
\mathcal{L}_{\mathrm{CTC}}(\widetilde{Z}_i^v,G_i).
\end{equation}

\textbf{Gloss-free SLT.}
Following MixSignGraph~\cite{gan2025mixsigngraph}, we train the frontal-view translation model with pseudo-gloss supervision. The pseudo glosses $\widetilde{G}_i$ are derived from the target sentence $S_i$ by punctuation removal and lemmatization, requiring no manual gloss annotations:
\begin{equation}
\mathcal{L}_{\mathrm{GFSLT}}^{\mathrm{front}}
=
\mathcal{L}_{\mathrm{CE}}(T_i^f, S_i)
+
\mathcal{L}_{\mathrm{CTC}}(Z_i^f,\widetilde{G}_i)
+
\mathcal{L}_{\mathrm{CTC}}(\widetilde{Z}_i^f,\widetilde{G}_i).
\end{equation}

During multi-view training, we supervise translation with the reference sentences using cross-entropy, and distill the translation logits from the frontal teacher at the corresponding target-token positions under teacher forcing, where both decoders receive the same ground-truth target prefix:
\begin{equation}
\mathcal{L}_{\mathrm{GFSLT}}^{\mathrm{multi}}
=
\mathcal{L}_{\mathrm{CE}}(T_i^v,S_i)
+
\lambda
\mathcal{D}_{\mathrm{KD}}(T_i^f,T_i^v),
\end{equation}
where $T_i^f$ denotes the pretrained translation logits from the frontal-view translation model.
Pseudo-gloss CTC provides auxiliary alignment during frontal-view pretraining; since pseudo glosses are automatically derived rather than manually annotated, multi-view training uses $\mathcal{L}_{\mathrm{CE}}$ and $\mathcal{D}_{\mathrm{KD}}$.

\section{Experiments}
\textbf{Experimental Settings.} All experiments are conducted on MVSign. MVSign-H2S is evaluated only for gloss-free SLT due to the absence of gloss annotations.
We use WER for CSLR task and BLEU-1--4 and ROUGE for SLT task. Detailed experimental settings are provided in Appendix~\ref{sec:supp_experimental_settings}.

\subsection{Discussion on Data Quality} 

\begin{wrapfigure}{r}{0.56\columnwidth}
  \vspace{-8mm}
  \centering
  \includegraphics[width=0.56\columnwidth]{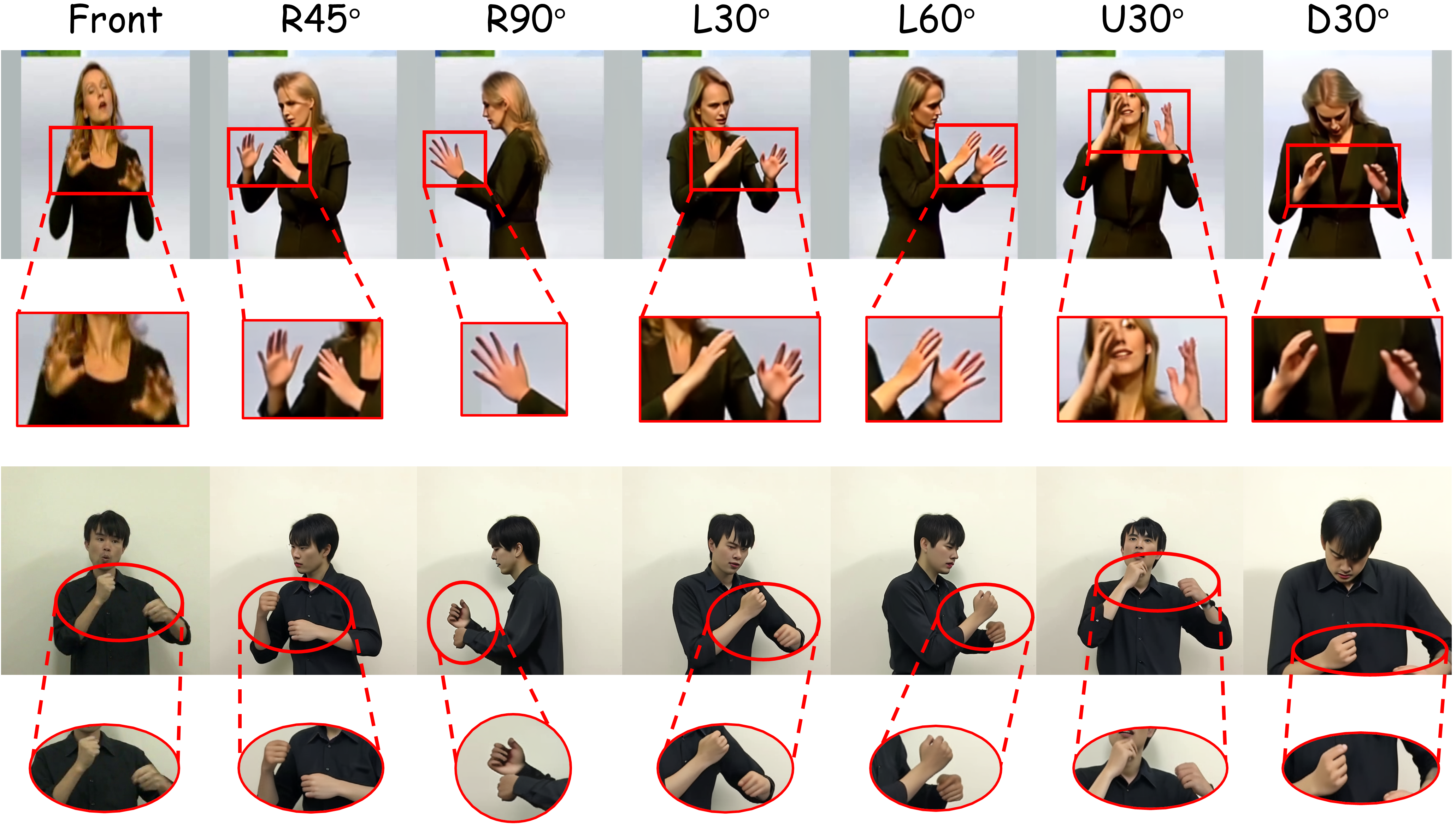}
  \vspace{-5mm}
  \caption{Visualization of the same sign sample under multiple viewpoints with evident appearance variations.
  }
  \label{fig:result1}
  \vspace{-4mm}
\end{wrapfigure}

\textbf{Qualitative Assessment of Visual Fidelity.}
Figure~\ref{fig:result1} shows two sign sequences from seven viewpoints.
The red annotations and enlarged crops highlight hand region, where viewpoint changes have a strong influence on visual appearance.
Across the views, the generated videos preserve coherent body contours and signer appearance while maintaining fine-grained hand details.
In particular, individual fingers remain visually distinguishable in non-frontal viewpoints and in configurations involving partial self-occlusion or closely grouped fingers, without obvious finger merging or structural artifacts.
These examples provide evidence that the rendering process preserves local visual cues relevant to sign recognition under viewpoint changes.

\begin{wraptable}{r}{0.6\textwidth}
\renewcommand\arraystretch{1.1}
\vspace{-3mm}
\caption{Pose fidelity in MVSign.}
\vspace{-3mm}
\label{tab:Pose fidelity in MVSign.}
\centering
\resizebox{0.6\textwidth}{!}{
\begin{tabular}{ccccc}
\toprule
    &\textbf{NME}{$\downarrow$} &\textbf{PCK@0.05}{$\uparrow$} &\textbf{PCK@0.10}{$\uparrow$} &\textbf{TJ-Err}{$\downarrow$} \\
\midrule
   Body &0.0132 &0.9814 &0.9962 &0.0075 \\
   Left Hand &0.0233 &0.9438 &0.9835 &0.0210 \\
   Right Hand &0.0139 &0.9844 &0.9971 &0.0164 \\
\bottomrule
\end{tabular}}
\vspace{-4mm}
\end{wraptable}

\textbf{Quantitative Pose-Fidelity Evaluation.}
The synthesized videos are generated by rotating the recovered 3D body around the global root, thereby preserving local joint configurations and temporal motion.
We assess geometric fidelity by projecting the transformed 3D joints under each target-view and comparing these reference poses with 2D body and hand keypoints re-extracted from the rendered videos using DWPose.
Because the references are derived directly from the transformed 3D motion, this evaluation measures consistency between the target geometry and the rendered observations.
Averaged over the non-frontal viewpoints, the results in Table~\ref{tab:Pose fidelity in MVSign.} yield NME $\leq 0.0233$, PCK@0.05 $\geq 0.9438$, PCK@0.10 $\geq 0.9835$, and TJ-Err (Temporal Jitter Error) $\leq 0.0210$, supporting the fidelity of motion for multi-view SLU.

\begin{table*}[t]
\renewcommand\arraystretch{1.2}
\caption{Comparison of CSLR performance on the MVSign dataset.}
\label{tab:Comparison of CSLR performance on the MVSign dataset.}
\vspace{-3mm}
\centering
\resizebox{1\textwidth}{!}{
\begin{tabular}{l c@{\hskip 0.1pt} cccc c cccc}
\toprule
   \multirow{2}{*}{\textbf{Methods}} &~ &\multicolumn{4}{c}{\textbf{MVSign-PT14}} &~ &\multicolumn{4}{c}{\textbf{MVSign-CSL}}\\
   \cline{3-6}\cline{8-11}
    &~ &\textbf{Dev-WER$\downarrow$} &\textbf{Dev-del$\downarrow$/ins$\downarrow$} &\textbf{Test-WER$\downarrow$} &\textbf{Test-del$\downarrow$/ins$\downarrow$} &~ &\textbf{Dev-WER$\downarrow$} &\textbf{Dev-del$\downarrow$/ins$\downarrow$} &\textbf{Test-WER$\downarrow$} &\textbf{Test-del$\downarrow$/ins$\downarrow$}\\
\midrule
    SMKD~\citep{hao2021self} &~ &\cellcolor{lightgray}42.83 &\multicolumn{1}{c|}{17.61/2.06} &\cellcolor{lightgray}42.91 &16.06/2.68 &~ &\cellcolor{lightgray}45.78 &\multicolumn{1}{c|}{16.56/2.23} &\cellcolor{lightgray}45.01 &16.24/2.24\\
    VAC~\citep{min2021visual} &~ &\cellcolor{lightgray}39.85 &\multicolumn{1}{c|}{16.20/2.35} &\cellcolor{lightgray}39.57 &16.21/3.21 &~ &\cellcolor{lightgray}46.62 &\multicolumn{1}{c|}{18.31/2.41} &\cellcolor{lightgray}44.59 &17.15/2.46 \\
    MMTLB~\citep{chen2022simple} &~ &\cellcolor{lightgray}38.89 &\multicolumn{1}{c|}{12.88/4.33} &\cellcolor{lightgray}38.05 &11.55/4.57 &~ &\cellcolor{lightgray}44.28 &\multicolumn{1}{c|}{16.44/3.71} &\cellcolor{lightgray}43.28 &15.02/3.74 \\
    C$^2$SLR~\citep{zuo2022c2slr} &~ &\cellcolor{lightgray}41.90 &\multicolumn{1}{c|}{19.10/2.40} &\cellcolor{lightgray}40.50 &18.50/2.9 &~ &\cellcolor{lightgray}46.64 &\multicolumn{1}{c|}{18.31/1.96} &\cellcolor{lightgray}45.22 &18.08/1.89 \\
    SEN~\citep{hu2023self} &~ &\cellcolor{lightgray}37.77 &\multicolumn{1}{c|}{14.77/2.61} &\cellcolor{lightgray}38.17 &14.18/3.34 &~ &\cellcolor{lightgray}42.23 &\multicolumn{1}{c|}{13.08/3.36} &\cellcolor{lightgray}41.11 &12.25/3.31 \\
    CorrNet~\citep{hu2023continuous} &~ &\cellcolor{lightgray}34.93 &\multicolumn{1}{c|}{12.50/2.36} &\cellcolor{lightgray}34.80 &11.61/2.95 &~ &\cellcolor{lightgray}41.21 &\multicolumn{1}{c|}{13.47/3.41} &\cellcolor{lightgray}40.21 &12.54/2.95\\
    SignGraph~\citep{gan2024signgraph} &~ &\cellcolor{lightgray}34.65 &\multicolumn{1}{c|}{11.58/2.65} &\cellcolor{lightgray}35.13 &10.52/3.53 &~ &\cellcolor{lightgray}38.63 &\multicolumn{1}{c|}{13.50/2.93} &\cellcolor{lightgray}37.63 &12.79/2.59\\
    HSTE~\citep{xu2025hierarchical} &~ &\cellcolor{lightgray}42.03 &\multicolumn{1}{c|}{17.21/2.11} &\cellcolor{lightgray}40.31 &18.23/2.95 &~ &\cellcolor{lightgray}44.33 &\multicolumn{1}{c|}{16.32/3.73} &\cellcolor{lightgray}42.74 &14.23/3.55\\   
    MSKA~\citep{guan2025mska} &~ &\cellcolor{lightgray}35.25 &\multicolumn{1}{c|}{11.96/2.51} &\cellcolor{lightgray}35.45 &10.78/3.10 &~ &\cellcolor{lightgray}38.07 &\multicolumn{1}{c|}{14.08/2.73} &\cellcolor{lightgray}37.39 &12.99/2.47\\
\midrule
    CanonSLU (Ours) &~ &\cellcolor{lightgray}\textbf{33.27} &\multicolumn{1}{c|}{12.95/2.25} &\cellcolor{lightgray}\textbf{33.35} &11.85/2.89 &~ &\cellcolor{lightgray}\textbf{36.92} &14.53/2.39 &\cellcolor{lightgray}\textbf{35.71} &13.68/2.27\\
\bottomrule
\end{tabular}}
\vspace{-3mm}
\end{table*}

\subsection{Comparison with State-of-the-Art Methods on CSLR Task}
Table~\ref{tab:Comparison of CSLR performance on the MVSign dataset.} compares representative state-of-the-art CSLR methods on the proposed MVSign-PT14 and MVSign-CSL subsets. For fair comparison, all competing methods are retrained on these two multi-view datasets following the training protocols and implementation details described in their original papers. \textbf{CanonSLU} achieves the lowest WER on all evaluation splits.

\textbf{MVSign-PT14.}
CanonSLU obtains WERs of 33.27\% and 33.35\% on the dev and test sets, respectively. 
Compared with the strongest baseline, SignGraph (35.13\%) and CorrNet (34.80\%) on the test set, CanonSLU reduces WER by \textbf{1.78} and \textbf{1.45}, respectively.
These results demonstrate the consistent effectiveness of CanonSLU for CSLR under multi-view settings.

\textbf{MVSign-CSL.}
CanonSLU obtains WERs of 36.92\% and 35.71\% on the dev and test sets, respectively. MSKA provides the strongest baseline performance on both splits, while CanonSLU reduces WER by \textbf{1.15} and \textbf{1.68} on the dev and test sets, respectively. These results further confirm the effectiveness of CanonSLU for robust multi-view CSLR.

\begin{table*}[t]
\centering
\caption{Comparison of SLT performance on the MVSign dataset.}
\label{tab:Comparison of SLT performance on the MVSign dataset.}
\vspace{-3mm}
\renewcommand{\arraystretch}{1.3}
\resizebox{\textwidth}{!}{
\begin{tabular}{clccccc ccccc}
\toprule
\multirow{2}{*}{\textbf{Task}} &\multirow{2}{*}{\makecell[l]{\textbf{Methods}}} &\multicolumn{5}{c}{\textbf{DEV}} &\multicolumn{5}{c}{\textbf{TEST}} \\
\cmidrule(l){3-7}
\cmidrule(l){8-12}
& &\makecell{\textbf{BLEU1$\uparrow$}} &\makecell{\textbf{BLEU2$\uparrow$}} &\makecell{\textbf{BLEU3$\uparrow$}} &\makecell{\textbf{BLEU4$\uparrow$}} &\makecell{\textbf{ROUGE$\uparrow$}} &\makecell{\textbf{BLEU1$\uparrow$}} &\makecell{\textbf{BLEU2$\uparrow$}} &\makecell{\textbf{BLEU3$\uparrow$}} &\makecell{\textbf{BLEU4$\uparrow$}} &\makecell{\textbf{ROUGE$\uparrow$}} \\
\midrule
\rowcolor{green!10}\multicolumn{12}{c}{\bfseries\emph{MVSign-PT14}} \\
\multirow{5}{*}{\rotatebox[origin=c]{90}{\textbf{Gloss-based}}}
&MMTLB~\citep{chen2022simple} 
&44.55 &31.56 &24.20 &19.55 &\multicolumn{1}{c|}{44.49}
&43.34 &30.84 &23.60 &19.04 &43.76 \\
&SLTUNET~\citep{zhang2023sltunet}
&45.61 &33.50 &25.83 &20.90 &\multicolumn{1}{c|}{46.01}
&43.96 &30.85 &23.91 &18.97 &44.79 \\
&CV-SLT~\citep{zhao2024conditional}
&45.36 &32.45 &25.79 &20.80 &\multicolumn{1}{c|}{46.67}
&44.06 &30.02 &23.55 &18.68 &44.38 \\
&MSKA~\citep{guan2025mska}
&44.78 &31.45 &23.85 &18.96 &\multicolumn{1}{c|}{43.51}
&43.98 &30.93 &23.31 &18.35 &42.40 \\
&\cellcolor{lightgray}\textbf{CanonSLU (Ours)}
&\cellcolor{lightgray}\textbf{46.66} &\cellcolor{lightgray}\textbf{33.68} &\cellcolor{lightgray}\textbf{26.11} &\cellcolor{lightgray}\textbf{21.19} &\multicolumn{1}{c|}{\cellcolor{lightgray}\textbf{46.89}}
&\cellcolor{lightgray}\textbf{44.54} &\cellcolor{lightgray}\textbf{31.98} &\cellcolor{lightgray}\textbf{24.53} &\cellcolor{lightgray}\textbf{19.73} &\cellcolor{lightgray}\textbf{45.47} \\
\multirow{5}{*}{\rotatebox[origin=c]{90}{\textbf{Gloss-free}}}
&GFSLT-VLP~\citep{zhou2023gloss} &31.67 &22.31 &17.09 &13.78 &\multicolumn{1}{c|}{31.83} &28.90 &19.80 &14.71 &11.75 &29.62\\
&Sign2GPT~\citep{wong2024sign2gpt} &35.61 &24.60 &18.34 &14.41 &\multicolumn{1}{c|}{35.30} &33.94 &23.62 &17.55 &13.84 &34.20\\
&SpaMo~\citep{hwang2025efficient} &36.30 &23.24 &17.05 &13.52 &\multicolumn{1}{c|}{30.92} &35.50 &22.70 &16.44 &12.99 &29.18\\
&MMSLT~\citep{kim2025leveraging} &37.40 &26.46 &20.13 &15.06 &\multicolumn{1}{c|}{36.11} &35.95 &25.54 &19.46 &14.68 &34.84\\
&\cellcolor{lightgray}\textbf{CanonSLU (Ours)} &\cellcolor{lightgray}\textbf{41.18} &\cellcolor{lightgray}\textbf{27.84} &\cellcolor{lightgray}\textbf{20.32} &\cellcolor{lightgray}\textbf{15.66} &\multicolumn{1}{c|}{\cellcolor{lightgray}\textbf{42.27}} &\cellcolor{lightgray}\textbf{39.48} &\cellcolor{lightgray}\textbf{26.85} &\cellcolor{lightgray}\textbf{19.78} &\cellcolor{lightgray}\textbf{15.33} &\cellcolor{lightgray}\textbf{40.77} \\
\midrule

\rowcolor{green!10}\multicolumn{12}{c}{\bfseries\emph{MVSign-CSL}} \\
\multirow{5}{*}{\rotatebox[origin=c]{90}{\textbf{Gloss-based}}}
&MMTLB~\citep{chen2022simple} &42.75 &30.11 &21.75 &16.29 &\multicolumn{1}{c|}{40.65} &42.65 &30.08 &21.79 &16.35 &40.74\\
&SLTUNET~\citep{zhang2023sltunet} &40.67 &28.32 &20.62 &15.55 &\multicolumn{1}{c|}{39.63} &40.94 &28.59 &20.80 &15.69 &40.57\\
&CV-SLT~\citep{zhao2024conditional} &41.32 &29.44 &20.58 &16.01 &\multicolumn{1}{c|}{40.73} &42.66 &29.82 &21.27 &16.57 &41.10\\
&MSKA~\citep{guan2025mska} &42.00 &28.48 &21.77 &16.07 &\multicolumn{1}{c|}{40.26} &42.37 &28.13 &21.47 &16.74 &40.34\\
&\cellcolor{lightgray}\textbf{CanonSLU (Ours)} &\cellcolor{lightgray}\textbf{43.46} &\cellcolor{lightgray}\textbf{30.75} &\cellcolor{lightgray}\textbf{22.76} &\cellcolor{lightgray}\textbf{17.47} &\multicolumn{1}{c|}{\cellcolor{lightgray}\textbf{41.42}} &\cellcolor{lightgray}\textbf{44.43} &\cellcolor{lightgray}\textbf{31.60} &\cellcolor{lightgray}\textbf{23.51} &\cellcolor{lightgray}\textbf{18.12} &\cellcolor{lightgray}\textbf{42.04} \\
\multirow{5}{*}{\rotatebox[origin=c]{90}{\textbf{Gloss-free}}}
&GFSLT-VLP~\citep{zhou2023gloss} &17.67 &7.72 &4.49 &3.06 &\multicolumn{1}{c|}{18.15} &17.14 &7.16 &4.00 &2.58 &17.57\\
&Sign2GPT~\citep{wong2024sign2gpt} &35.38 &22.96 &14.88 &9.87 &\multicolumn{1}{c|}{33.84} &35.96 &23.28 &14.99 &9.78 &34.21\\
&SpaMo~\citep{hwang2025efficient} &29.32 &18.07 &12.23 &8.74 &\multicolumn{1}{c|}{27.83} &29.60 &17.93 &12.13 &8.72 &27.06\\
&MMSLT~\citep{kim2025leveraging} &36.32 &23.13 &14.02 &10.38 &\multicolumn{1}{c|}{36.79} &35.94 &23.54 &14.65 &10.66 &36.72\\
&\cellcolor{lightgray}\textbf{CanonSLU (Ours)} &\cellcolor{lightgray}\textbf{39.91} &\cellcolor{lightgray}\textbf{27.22} &\cellcolor{lightgray}\textbf{19.67} &\cellcolor{lightgray}\textbf{14.90} &\multicolumn{1}{c|}{\cellcolor{lightgray}\textbf{38.04}} &\cellcolor{lightgray}\textbf{40.93} &\cellcolor{lightgray}\textbf{28.11} &\cellcolor{lightgray}\textbf{20.39} &\cellcolor{lightgray}\textbf{15.47} &\cellcolor{lightgray}\textbf{38.84} \\
\midrule

\rowcolor{green!10}\multicolumn{12}{c}{\bfseries\emph{MVSign-H2S}} \\
\multirow{4}{*}{\rotatebox[origin=c]{90}{\textbf{Gloss-free}}}
&GFSLT-VLP~\citep{zhou2023gloss} &14.76 &9.46 &4.83 &1.93 &\multicolumn{1}{c|}{13.37} &14.23 &8.98 &4.72 &2.06 &13.51 \\
&Sign2GPT~\citep{wong2024sign2gpt} &16.61 &11.32 &7.04 &4.55 &\multicolumn{1}{c|}{16.47} &16.70 &11.01 &7.11 &4.63 &16.93\\
&SpaMo~\citep{hwang2025efficient} &17.21 &12.38 &7.17 &4.24 &\multicolumn{1}{c|}{17.83} &17.06 &11.58 &7.74 &4.33 &17.99\\
&\cellcolor{lightgray}\textbf{CanonSLU (Ours)} &\cellcolor{lightgray}\textbf{20.32} &\cellcolor{lightgray}\textbf{14.87} &\cellcolor{lightgray}\textbf{9.61} &\cellcolor{lightgray}\textbf{6.19} &\multicolumn{1}{c|}{\cellcolor{lightgray}\textbf{20.71}} &\cellcolor{lightgray}\textbf{20.77} &\cellcolor{lightgray}\textbf{14.42} &\cellcolor{lightgray}\textbf{8.98} &\cellcolor{lightgray}\textbf{6.62} &\cellcolor{lightgray}\textbf{20.90} \\
\bottomrule
\end{tabular}
}
\vspace{-5mm}
\end{table*}

\subsection{Comparison with State-of-the-Art Methods on SLT Task}
Table~\ref{tab:Comparison of SLT performance on the MVSign dataset.} compares representative state-of-the-art SLT methods under gloss-based and gloss-free settings on MVSign. 
We follow the same evaluation protocol as in the CSLR experiments to ensure a fair comparison. 
\textbf{CanonSLU} achieves the best performance across all datasets and evaluation splits.

\textbf{MVSign-PT14.}
CanonSLU achieves the best performance under both gloss-based and gloss-free settings. With gloss supervision, it obtains BLEU-4 scores of 21.19 and 19.73 on the dev and test sets, respectively, and ranks first across all metrics. On the dev set, it surpasses MSKA by \textbf{2.23} BLEU-4 and \textbf{3.38} ROUGE. Without gloss supervision, CanonSLU achieves ROUGE scores of 42.27 and 40.77 on the dev and test sets, respectively, maintaining strong performance.

\textbf{MVSign-CSL.}
CanonSLU ranks first across all metrics under both supervision settings. With gloss supervision, it achieves BLEU-4/ROUGE scores of 18.12/42.04 on the test set, exceeding the strongest competing results by \textbf{1.38}/\textbf{0.94}, respectively. Without gloss supervision, it obtains BLEU-4/ROUGE scores of 15.47/38.84, improving over MMSLT by \textbf{4.81}/\textbf{2.12} on the test set, respectively. CanonSLU thus maintains its advantage under both supervision settings.

\textbf{MVSign-H2S.}
CanonSLU ranks first across all metrics, achieving BLEU-4/ROUGE scores of 6.19/20.71 on the dev set and 6.62/20.90 on the test set. Compared with the SpaMo, CanonSLU improves BLEU-4 by \textbf{1.95} and \textbf{2.29} and ROUGE by \textbf{2.88} and \textbf{2.91} on both sets, respectively. These results demonstrate its effectiveness on real-world multi-view data.

\subsection{Ablation Study}
\textbf{Effectiveness of the CKT and MRA.}
We evaluate the individual contributions of the proposed modules on MVSign-PT14 across different viewpoints, as summarized in Table~\ref{tab:Ablation results of different module combinations for CSLR on the MVSign-PT14.}.
Here, ``Avg'' denotes the mean WER of two views in each category, and $\Delta_{\mathrm{Avg}}$ denotes the improvement relative to the baseline. 
The baseline model degrades significantly under large viewpoint variations, particularly for pitch views (\emph{D30$^\circ$}, \emph{U30$^\circ$}) and extreme horizontal angles (\emph{R90$^\circ$}, \emph{L60$^\circ$}). 
Introducing \textbf{CKT} reduces the overall WER by \textbf{2.25}, with larger gains on challenging views, including \textbf{1.62} for large-angle views and \textbf{4.05} for pitch views. This indicates that canonical-view supervision effectively alleviates projection-induced semantic ambiguity.
Applying \textbf{MRA} alone yields a larger overall gain of \textbf{2.86} and improves every viewpoint, indicating that the proposed MRA configuration benefits multi-view recognition.
Combining both modules achieves the best overall WER of \textbf{33.35\%}. 
The improvement is especially pronounced under extreme viewpoint changes, where pitch views achieve \textbf{8.38} reduction. 
These results indicate that CKT and MRA are complementary, addressing cross-view discrepancy at the semantic and representation levels, respectively.
Although CKT is applied only to non-frontal samples, its gradients update the model parameters shared across views, which may also improve frontal-view performance.

\begin{table*}[t]
\renewcommand\arraystretch{1.2}
\caption{Ablation results of different module combinations for CSLR on the MVSign-PT14. ``All'' represents all data; ``R'', ``L'', ``D'', and ``U'' represent right rotation, left rotation, downward tilt and upward tilt, respectively. $\Delta$ indicates the performance change compared with the Baseline. 
}
\label{tab:Ablation results of different module combinations for CSLR on the MVSign-PT14.}
\vspace{-3mm}
\centering
\resizebox{\textwidth}{!}{
\begin{tabular}{ccc| cc| ccc| ccc| ccc| cc}
\toprule
   \multicolumn{3}{c|}{\textbf{Methods}} &\multicolumn{2}{c|}{\textbf{Multi-view}} &\multicolumn{3}{c|}{\textbf{Large angle}} &\multicolumn{3}{c|}{\textbf{Small angle}} &\multicolumn{3}{c|}{\textbf{Pitch}} &\multicolumn{2}{c}{\textbf{Front}}\\
    \textbf{Baseline} &\textbf{CKT} &\textbf{MRA} &\textbf{All} &\textbf{$\Delta$} &\textbf{R90$^\circ$} &\textbf{L60$^\circ$} &\textbf{$\Delta_{\text{Avg}}$} &\textbf{R45$^\circ$} &\textbf{L30$^\circ$} &\textbf{$\Delta_{\text{Avg}}$} &\textbf{D30$^\circ$} &\textbf{U30$^\circ$} &\textbf{$\Delta_{\text{Avg}}$} &\textbf{0$^\circ$} &\textbf{$\Delta$}\\
\midrule
    \ding{51} && &37.73 &\cellcolor{lightgray}- &58.04 &50.73 &\cellcolor{lightgray}- &53.91 &52.52 &\cellcolor{lightgray}- &55.05 &52.10 &\cellcolor{lightgray}- &22.99 &\cellcolor{lightgray}-\\
    
    \ding{51} &\ding{51} & &35.48 &\cellcolor{lightgray}$\downarrow$2.25 &57.58 &47.95 &\cellcolor{lightgray}$\downarrow$1.62 &50.23 &52.52 &\cellcolor{lightgray}$\downarrow$1.84 &50.23 &48.83 &\cellcolor{lightgray}$\downarrow$4.05 &20.99 &\cellcolor{lightgray}$\downarrow$2.00\\
    
    \ding{51} & &\ding{51} &34.87 &\cellcolor{lightgray}$\downarrow$2.86 &55.13 &46.16 &\cellcolor{lightgray}$\downarrow$3.74 &49.46 &48.74 &\cellcolor{lightgray}$\downarrow$4.12 &44.34 &46.72 &\cellcolor{lightgray}$\downarrow$8.05 &22.70 &\cellcolor{lightgray}$\downarrow$0.29\\
   \ding{51} &\ding{51} &\ding{51}  &\textbf{33.35} &\cellcolor{lightgray}\textbf{$\downarrow$4.38} &\textbf{52.37} &\textbf{45.31} &\cellcolor{lightgray}$\downarrow$\textbf{5.55} &\textbf{47.01} &\textbf{47.95} &\cellcolor{lightgray}\textbf{$\downarrow$5.74} &\textbf{43.74} &\textbf{46.66} &\cellcolor{lightgray}\textbf{$\downarrow$8.38} &\textbf{20.66} &\cellcolor{lightgray}\textbf{$\downarrow$2.33}\\
\bottomrule
\end{tabular}}
\end{table*}

\begin{figure*}[t]
  \centering
  \includegraphics[width=\linewidth]{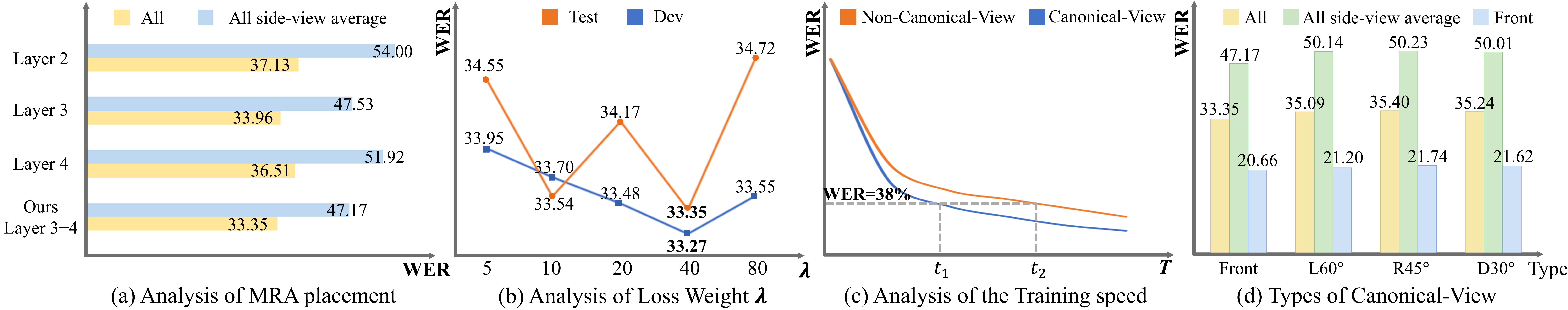}
  \vspace{-6mm}
  \caption{Detailed analysis of the designed CanonSLU for CSLR on the MVSign-PT14, where (a), (c), (d) report the performance on the test set. (a) Hierarchical Placement Analysis of the MRA Module. (b) Sensitivity analysis of the distillation weight $\lambda$. (c) Comparison of canonical-view and non-canonical-view guidance. (d) Comparison of different canonical-view anchors.}
  \label{fig:analysis}
  \vspace{-3mm}
\end{figure*}

\textbf{Mid-to-high layers are optimal for MRA.}
We analyze the impact of inserting MRA at different encoder stages, as shown in Fig.~\ref{fig:analysis}(a). 
The \texttt{layer1} setting is omitted due to prohibitive memory cost. 
Among single-layer configurations, \texttt{layer3} achieves the best performance, indicating that mid-to-high-level features are more suitable for modeling stable temporal relations. 
Extending MRA to \texttt{layer3+4} further improves non-frontal viewpoints performance, suggesting that motion representations from \texttt{layer3} and semantic representations from \texttt{layer4} are complementary. 
Considering both performance and computational cost, we adopt \texttt{layer3+4} in the final model.

\textbf{Moderate $\lambda$ gives the best trade-off.}
To evaluate the robustness of the proposed CKT, we conduct a sensitivity analysis on the distillation weight $\lambda$, varying it over $\{5, 10, 20, 40, 80\}$ (Figure~\ref{fig:analysis}(b)). 
The dev curve exhibits a relatively stable descent, reaching its optimal performance of 33.27\% at $\lambda=40$. 
When $\lambda=80$, the performance slightly degrades, suggesting that an excessively large distillation weight may overwhelm the hard-label CTC constraints and cause the model to overfit the canonical-view. 
While the test set demonstrates more fluctuation—a common phenomenon under complex multi-view settings—it perfectly aligns with the dev set at the optimal threshold, yielding the lowest test WER of 33.35\% at $\lambda=40$. 
Consequently, we set $\lambda=40$ as the default configuration, ensuring an optimal equilibrium between task-specific learning and cross-view semantic transfer.

\textbf{Canonical-view guidance improves optimization.}
To assess the effect of canonical-view prior guidance, we compare training with and without it. 
As shown in Figure~\ref{fig:analysis}(c), the guidance accelerates convergence, reaching the same WER earlier ($t_2 > t_1$), while maintaining consistently lower error rates throughout training.
This suggests that the canonical view provides more stable supervision, improving both optimization efficiency and final recognition performance.

\textbf{The frontal view is the most effective canonical anchor.}
We further analyze which view is the most suitable choice as the canonical anchor: \emph{Front}, \emph{L60$^\circ$}, \emph{R45$^\circ$}, and \emph{D30$^\circ$}. 
All canonical-anchor models are trained using the same number of samples.
The results are summarized in Figure~\ref{fig:analysis}(d), we report the WER on the \textbf{All}, the \textbf{Front}, and the average WER over \textbf{All side-view}. 
Among all views, using the frontal view as the canonical anchor yields the lowest WER (\emph{e.g.} 33.35\% (All), 47.17\% (All side-view) and 20.66\% (Front)). 
These results suggest that the frontal view serves as the most effective canonical anchor, as it provides the most complete and stable visual knowledge.

\begin{wraptable}{r}{0.61\textwidth}
\renewcommand\arraystretch{1.2}
\vspace{-3mm}
\caption{Accuracy-efficiency comparison on MVSign-PT14.}
\label{tab:Accuracy-efficiency comparison on MVSign-PT14.}
\vspace{-3mm}
\centering
\resizebox{0.61\textwidth}{!}{
\begin{tabular}{lcccc}
\toprule
    \textbf{Methods}& \textbf{WER{$\downarrow$}} & \textbf{Params(M){$\downarrow$}} & \textbf{FLOPs(G){$\downarrow$}} & \textbf{Inference Speed(videos/s){$\uparrow$}} \\
\midrule
    VAC &39.57 &\underline{34.30} &\textbf{417.57} &\textbf{44.51} \\
    SEN &38.17 &34.51 &423.15 &26.01 \\
    CorrNet &\underline{34.80} &\textbf{33.37} &427.97 &26.03 \\
    SignGraph &35.13 &36.63 &491.12 &17.30 \\
    \cellcolor{lightgray}CanonSLU &\cellcolor{lightgray}\textbf{33.35} &\cellcolor{lightgray}35.35 &\cellcolor{lightgray}\underline{420.83} &\cellcolor{lightgray}\underline{36.35} \\
\bottomrule
\end{tabular}}
\vspace{-3mm}
\end{wraptable}

\textbf{Performance-complexity analysis.}
Data generation is used only for offline dataset construction and does not affect inference.
All experiments are conducted on the same server under the same GPU, batch-size, input-resolution, numerical-precision, and evaluation settings.
As shown in Table~\ref{tab:Accuracy-efficiency comparison on MVSign-PT14.}, CanonSLU achieves the lowest WER with competitive complexity, outperforming CorrNet in inference speed and SignGraph in parameter count, FLOPs, and speed, thereby offering a favorable accuracy--efficiency trade-off.

\begin{wrapfigure}{r}{0.5\columnwidth}
  \vspace{-4mm}
  \centering
  \includegraphics[width=0.5\columnwidth]{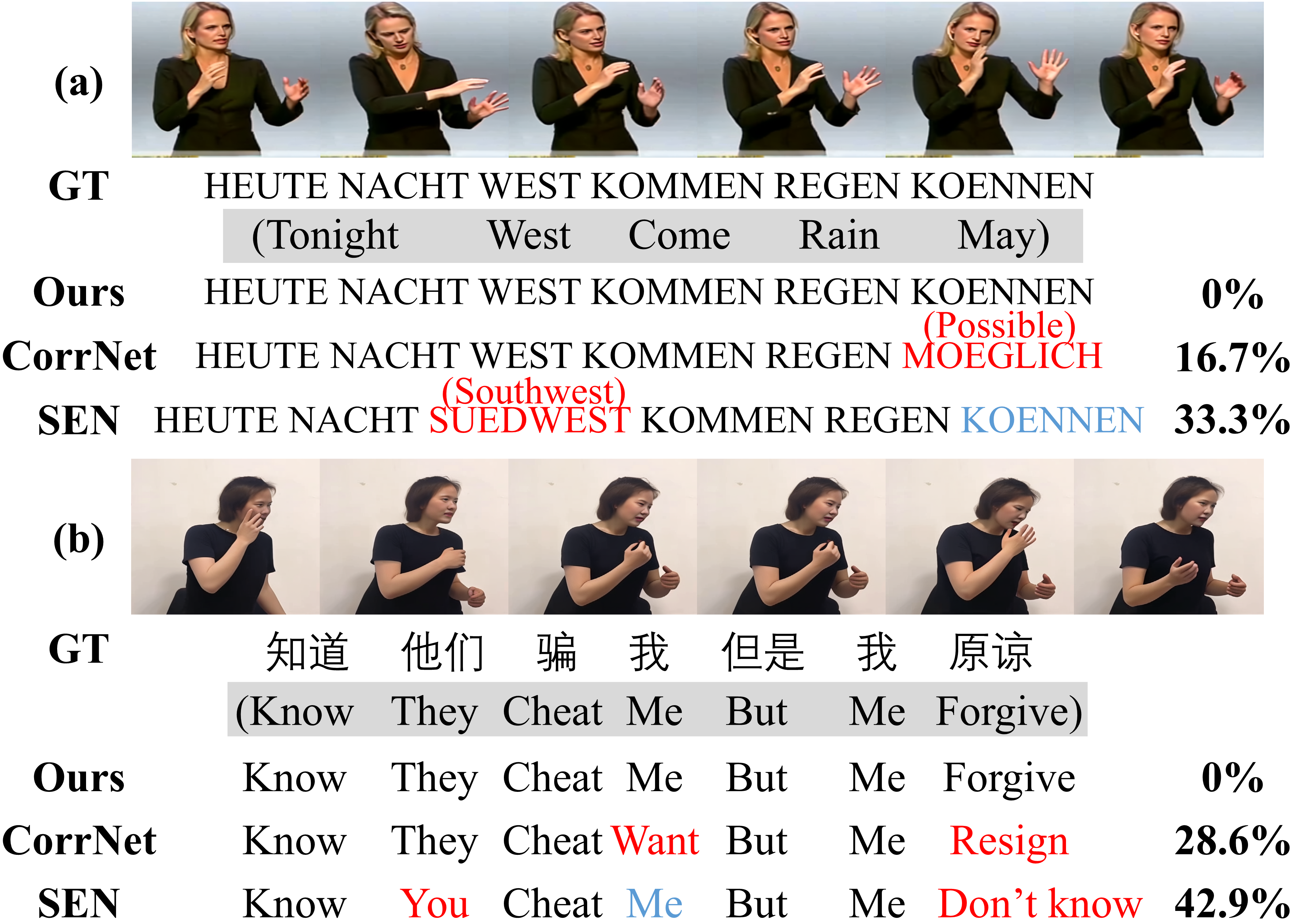}
  \vspace{-4mm}
  \caption{Qualitative comparison of different methods. Red words indicate incorrect predictions, while blue words denote missing glosses.}
  \label{fig:result}
  \vspace{-3mm}
\end{wrapfigure}

\textbf{Visualization of Recognition Results.}
As shown in Figure~\ref{fig:result}, we further present a qualitative comparison of different methods. Under challenging viewpoint conditions, existing methods are more prone to semantic confusion and missing glosses. For example, in case (a), CorrNet misrecognizes ``KOENNEN'' as ``MOEGLICH'', while SEN further confuses ``WEST'' with ``SUEDWEST''; in case (b), both CorrNet and SEN produce more obvious word substitution and semantic deviation errors. In contrast, our method correctly recovers the complete gloss sequences, demonstrating stronger robustness under multi-view settings. This indicates that CKT effectively alleviates cross-view semantic discrepancy, while MRA further refines stable motion cues.

\section{Conclusions}
In this paper, we introduce \textbf{MVSign}, a multi-view SLU benchmark spanning three sign languages and seven viewpoints, and propose \textbf{CanonSLU}, a canonical-view-guided baseline for multi-view SLU.
MVSign combines naturally captured recordings with additional views synthesized through 3D whole-body motion reconstruction and target-view rendering, while preserving the source splits and annotations.
CanonSLU treats the frontal view as a semantic anchor, combining CKT to transfer canonical knowledge to non-frontal views with MRA to strengthen motion-aware temporal representations under viewpoint changes. 
Experiments show that CanonSLU improves non-frontal performance, validating canonical-view guidance.
As a controlled benchmark, MVSign still requires future validation on real multi-camera data and may introduce a synthetic-to-real gap.

\subsection*{Reproducibility statement}
To ensure reproducibility, we provide the following resources:  All implementation details, including network architectures, hyperparameters, and training protocols, are described in Appendix~\ref{sub_Implementation_Details}. 
The CanonSLU implementation, data-generation scripts, and dataset metadata will be released upon publication.
Random seeds are fixed for all reported experiments.




\bibliography{references}
\bibliographystyle{iclr2027_conference}

\newpage

\appendix
\section*{\centering Supplementary Material}

\phantomsection

\noindent\textbf{Contents}\par
\medskip

\noindent
\hyperref[sec:supp_Related work]{A. Related Work}
\dotfill
\pageref{sec:supp_Related work}
\par

\noindent
\hyperref[sec:supp_mra]{B. Detailed Implementation of Motion Relation Aggregation}
\dotfill
\pageref{sec:supp_mra}
\par

\noindent
\hyperref[sec:supp_experimental_settings]{C. Experimental Settings}
\dotfill
\pageref{sec:supp_experimental_settings}
\par

\noindent
\hyperref[sec:supp_Additional Experimental Results]{D. Additional Experimental Results}
\dotfill
\pageref{sec:supp_Additional Experimental Results}
\par

\vspace{1em}

\section{Related work}
\label{sec:supp_Related work}
\subsection{Sign Language Understanding}
\textbf{CSLR.}
Early approaches rely on handcrafted features~\citep{freeman1995orientation,han2009modelling} and HMM-based temporal modeling~\citep{koller2015continuous,koller2017Re-sign}. Deep learning methods adopt CNN- and RNN-based architectures~\citep{koller2019weakly,guodan2019ctm,hao2021self,min2021visual}, typically combining 2D or 3D CNN backbones with temporal modules such as LSTMs~\citep{shi2015convolutional}. The CTC loss enables end-to-end training without frame-level gloss annotations~\citep{graves2006connectionist,hu2023continuous,gan2024signgraph,hu2023self}. Recent studies incorporate auxiliary cues, including skeletons~\citep{zhou2020spatial,yang2024s2net}, local regions~\citep{zuo2022c2slr,jiao2023cosign}, and depth information~\citep{jiang2021skeleton,wang2020fast}, while improving backbones, attention, contrastive learning, temporal modeling, and large-scale pre-training~\citep{cheng2020fully,gan2023contrastive,li2025Uni-Sign}.
Online CSLR extend recognition and translation to streaming inputs, addressing the latency and memory demands associated with processing complete videos~\citep{zuo2024towards}.
However, fine-grained hand-motion modeling remains challenging under viewpoint changes that alter hand appearance, motion projection, and occlusion.

\textbf{SLT.}
Early methods formulate SLT as sequence-to-sequence learning with intermediate gloss supervision~\citep{camgoz2018neural}. Later works unify gloss recognition and sentence generation~\citep{camgoz2020sign} and improve modeling with multi-stream architectures~\citep{chen2022two}, graph and contextual modeling~\citep{tang2021graph,jang2025lost}, and multimodal or visual pre-training~\citep{chen2022simple,jiao2024visual}. 
Recent work further investigates multilingual scaling and cross-modal transfer through joint pretraining on sign language translation and text translation tasks~\citep{zhang2024scaling}.
Gloss-free SLT avoids costly gloss annotations and explores visual-language pre-training~\citep{zhou2023gloss}, sign representation learning~\citep{yin2023gloss,gan2026learning}, pseudo-gloss and contrastive learning~\citep{gan2025mixsigngraph,guo2025bridging,lai2026selective}, and large language models~\citep{wong2024sign2gpt,kim2025leveraging}. However, most methods assume single or fixed views, leaving viewpoint robustness underexplored.

\subsection{Sign Language Datasets}
Large-scale, high-quality datasets are essential for deep SLU. Existing benchmarks target different tasks: WLASL~\citep{li2020word}, MM-WLAuslan~\citep{shen2024mm}, National CSL-DP~\citep{jing2025dual}, and Multi-VSL~\citep{dinh2025sign} support ISLR, with some providing multi-view samples, while PHOENIX-2014T~\citep{camgoz2018neural} and CSL-Daily~\citep{zhou2021improving} support CSLR and SLT with gloss and sentence annotations. How2Sign~\citep{duarte2021how2sign} provides multimodal, multiview continuous ASL videos with English transcripts but no gloss labels. Web-collected corpora, including BoBSL~\citep{albanie2021bbc}, JWSign~\citep{gueuwou2023jwsign}, YouTube-ASL~\citep{uthus2023youtube}, and YouTube-SL-25~\citep{tanzer2024youtube}, further expand dataset scale. Nevertheless, multi-view resources for continuous SLU remain scarce, as most CSLR and SLT benchmarks use fixed views and existing multi-view datasets mainly target ISLR. Since real multi-view collection requires calibrated and synchronized cameras with aligned annotations, we construct semantically aligned multi-view data from single-view videos through 3D reconstruction and view rendering.

\section{Detailed Implementation of Motion Relation Aggregation}
\label{sec:supp_mra}

Given an intermediate feature map
$A_{i,l}^{v}\in\mathbb{R}^{C_l\times T_i\times H_l\times W_l}$,
we first compute the frame-wise difference representation:
\begin{equation}
M_{i,l}^{v}(t)
=
A_{i,l}^{v}(t)-A_{i,l}^{v}(t-1),
\qquad t=2,\ldots,T_i.
\end{equation}
The first temporal position is initialized as
$M_{i,l}^{v}(1)=M_{i,l}^{v}(2)$.
The difference representation is then refined by a temporal convolution:
\begin{equation}
\widetilde{M}_{i,l}^{v}
=
\phi_m(M_{i,l}^{v}),
\end{equation}
where $\phi_m(\cdot)$ consists of a $3\times1\times1$ temporal
convolution, batch normalization, and ReLU activation.
In parallel, the original feature is projected into a value representation:
\begin{equation}
V_{i,l}^{v}
=
\phi_v(A_{i,l}^{v}),
\end{equation}
where $\phi_v(\cdot)$ is a $1\times1\times1$ convolution followed
by batch normalization.

For batched aggregation, we omit the sample, stage, and view indices
and write the feature shape as $B\times C\times T\times H\times W$,
where $B$ is the batch size.
We flatten the spatial dimensions into $N=HW$ locations.
For each valid temporal offset
$\delta\in\mathcal{T}_{\mathrm{off}}=\{1,2,4\}$ with $\delta<T$,
we pair the earlier frames $1,\ldots,T-\delta$ with the later
frames $1+\delta,\ldots,T$.
The corresponding refined motion features are denoted by
$\widetilde{M}^{-}$ and $\widetilde{M}^{+}$, and their value
features by $V^{-}$ and $V^{+}$.
For each pair index $t=1,\ldots,T-\delta$, channel normalization gives:
\begin{equation}
\overline{M}^{-}_{b,t,n}
=
\frac{\widetilde{M}^{-}_{b,:,t,n}}
{\max\left(\left\lVert\widetilde{M}^{-}_{b,:,t,n}\right\rVert_2,\varepsilon\right)},
\qquad
\overline{M}^{+}_{b,t,m}
=
\frac{\widetilde{M}^{+}_{b,:,t,m}}
{\max\left(\left\lVert\widetilde{M}^{+}_{b,:,t,m}\right\rVert_2,\varepsilon\right)}.
\end{equation}

Here, $\varepsilon>0$ prevents division by zero, $b$ indexes
the batch, and $n,m\in\{1,\ldots,N\}$ index
spatial locations in the earlier and later frames, respectively.
The motion similarity is computed by cosine similarity:
\begin{equation}
S_{b,t,n,m}
=
\left(\overline{M}^{-}_{b,t,n}\right)^{\mathsf{T}}
\overline{M}^{+}_{b,t,m}.
\end{equation}

For each earlier-frame location $n$, we select the top-$k$
later-frame locations according to $S_{b,t,n,m}$.
Let $\mathcal{N}_{k}^{-\rightarrow+}(n)$ denote this neighborhood,
with its dependence on $b$, $t$, and $\delta$ omitted for brevity.
The selected similarity scores are normalized as:
\begin{equation}
w_{b,t,n,m}^{-\rightarrow+}
=
\frac{\exp(S_{b,t,n,m})}
{\sum_{q\in\mathcal{N}_{k}^{-\rightarrow+}(n)}
\exp(S_{b,t,n,q})},
\qquad
m\in\mathcal{N}_{k}^{-\rightarrow+}(n).
\end{equation}
The selected value features are aggregated as:
\begin{equation}
O_{b,:,t,n}^{-\rightarrow+}
=
\sum_{m\in\mathcal{N}_{k}^{-\rightarrow+}(n)}
w_{b,t,n,m}^{-\rightarrow+}
V^{+}_{b,:,t,m}.
\end{equation}
The superscript $-\rightarrow+$ denotes matching an earlier-frame
location against later-frame locations.
The resulting message is received by the earlier-frame location.

For the reverse matching direction, we exchange the earlier and
later frames. For each later-frame location $m$, we independently
select the top-$k$ earlier-frame locations using $S_{b,t,n,m}$
over $n$ and apply softmax over the selected locations.
This gives $\mathcal{N}_{k}^{+\rightarrow-}(m)$ and
$w_{b,t,m,n}^{+\rightarrow-}$, respectively.
The reverse aggregation is:
\begin{equation}
O_{b,:,t,m}^{+\rightarrow-}
=
\sum_{n\in\mathcal{N}_{k}^{+\rightarrow-}(m)}
w_{b,t,m,n}^{+\rightarrow-}
V^{-}_{b,:,t,n}.
\end{equation}
This message is received by the later-frame location.

For each offset $\delta$, the messages
$O^{-\rightarrow+}$ and $O^{+\rightarrow-}$ are accumulated
at temporal positions $1,\ldots,T-\delta$ and
$1+\delta,\ldots,T$, respectively.
Let $c_t$ denote the number of messages received at temporal
position $t$ across all valid offsets.
The accumulated feature at each position is divided by
$\max(c_t,1)$, with the denominator broadcast over the batch,
channel, and spatial dimensions.
After restoring the spatial dimensions, the averaged feature
is denoted by $O_{i,l}^{v}$.

The aggregated feature is transformed by an output temporal convolution:
\begin{equation}
\widetilde{O}_{i,l}^{v}
=
\phi_o(O_{i,l}^{v}),
\end{equation}
where $\phi_o(\cdot)$ consists of a $3\times1\times1$ temporal
convolution and batch normalization.
The gate is computed from the original feature and the
nonnegative refined motion feature, both in their original
spatial shape:
\begin{equation}
G_{i,l}^{v}
=
\operatorname{sigmoid}
\left(
\phi_g\left(
[A_{i,l}^{v};\widetilde{M}_{i,l}^{v}]
\right)
\right).
\end{equation}
Here, $\phi_g(\cdot)$ is a $1\times1\times1$ convolution,
and $[\,;\,]$ denotes channel-wise concatenation.
The MRA output is:
\begin{equation}
Y_{i,l}^{v}
=
\widetilde{O}_{i,l}^{v}\odot G_{i,l}^{v},
\end{equation}
where $\odot$ denotes element-wise multiplication.
It is injected into the encoder through the residual update:
\begin{equation}
\widehat{A}_{i,l}^{v}
=
A_{i,l}^{v}+\alpha_lY_{i,l}^{v},
\end{equation}
where $\alpha_l$ is a learnable weight.
We use $k=24$ and $k=12$ at stages 3 and 4, respectively.
Algorithm~\ref{alg:mra} summarizes the aggregation and gating operations.

\textbf{Computational complexity.}
For each valid temporal offset $\delta$, dense similarity
computation requires $O(B(T-\delta)CN^2)$ time and
$O(B(T-\delta)N^2)$ storage for the similarity tensor.
Aggregating the selected top-$k$ value features requires
$O(B(T-\delta)NkC)$ time.
These operations are performed in both matching directions
and repeated over the valid offsets, with additional overhead
from convolutional projections and top-$k$ selection.
Since selection follows dense similarity computation, the
matching cost remains quadratic in the number of spatial
positions $N$.
This motivates placing MRA at stages 3 and 4, where the
feature maps are smaller.

\begin{algorithm}[h]
\caption{Motion Relation Aggregation}
\label{alg:mra}
\begin{algorithmic}[1]
\Require Feature map
$A\in\mathbb{R}^{B\times C\times T\times H\times W}$ with $T\ge2$,
temporal offset set $\mathcal{T}_{\mathrm{off}}=\{1,2,4\}$,
and top-$k$ parameter $k$
\Ensure MRA output $Y$

\State Compute frame differences:
$M_t\leftarrow A_t-A_{t-1}$ for $t=2,\ldots,T$
\State Initialize the first temporal position: $M_1\leftarrow M_2$
\State Refine motion features: $\widetilde{M}\leftarrow\phi_m(M)$
\State Compute value features: $V\leftarrow\phi_v(A)$
\State Flatten the spatial dimensions of $\widetilde{M}$ and $V$,
with $N=HW$
\State Initialize aggregated messages:
$O\leftarrow\mathbf{0}\in\mathbb{R}^{B\times C\times T\times N}$
\State Initialize per-position counts:
$c\leftarrow\mathbf{0}\in\mathbb{R}^{T}$

\ForAll{$\delta\in\mathcal{T}_{\mathrm{off}}$}
    \If{$\delta<T$}
        \State Form temporally offset pairs
        $(\widetilde{M}^{-},\widetilde{M}^{+})$ and $(V^{-},V^{+})$
        \State Normalize $\widetilde{M}^{-}$ and $\widetilde{M}^{+}$
        along the channel dimension with denominator
        $\max(\lVert\cdot\rVert_2,\varepsilon)$
        \State Compute dense cosine similarities across each frame pair
        \State Independently select top-$k$ matches and apply softmax
        in both matching directions
        \State Aggregate $V^{+}$ into the earlier positions
        $1,\ldots,T-\delta$
        \State Aggregate $V^{-}$ into the later positions
        $1+\delta,\ldots,T$
        \State Accumulate both sets of messages into $O$
        \State $c_{1:T-\delta}\leftarrow c_{1:T-\delta}+1$
        \State $c_{1+\delta:T}\leftarrow c_{1+\delta:T}+1$
    \EndIf
\EndFor

\State Normalize each temporal position:
$O_{b,:,t,:}\leftarrow O_{b,:,t,:}/\max(c_t,1)$
\State Reshape $O$ and $\widetilde{M}$ to
$\mathbb{R}^{B\times C\times T\times H\times W}$
\State Apply the output projection: $\widetilde{O}\leftarrow\phi_o(O)$
\State Compute the gate:
$G\leftarrow\operatorname{sigmoid}
\bigl(\phi_g([A;\widetilde{M}])\bigr)$
\State Modulate the aggregated feature: $Y\leftarrow\widetilde{O}\odot G$
\State \Return $Y$
\end{algorithmic}
\end{algorithm}

\section{Experimental Settings}
\label{sec:supp_experimental_settings}

\subsection{Datasets.}
All experiments are conducted on \textbf{MVSign}, which consists of \textbf{MVSign-PT14}, \textbf{MVSign-CSL}, and \textbf{MVSign-H2S}. 
MVSign-PT14 and MVSign-CSL are derived from PHOENIX-2014T~\citep{camgoz2018neural} and CSL-Daily~\citep{zhou2021improving}, respectively. For each subset, we retain all original videos as the \emph{Front} view and independently sample 800, 100, and 100 clips from the original train, dev, and test splits for each of the six non-frontal viewpoints $\{\mathrm{R45}^{\circ},\mathrm{R90}^{\circ},\mathrm{L30}^{\circ},\mathrm{L60}^{\circ},\mathrm{U30}^{\circ},\mathrm{D30}^{\circ}\}$. 
The sampled clips are converted into the corresponding target views using the construction process described in Section~\ref{sec:mvsign}, yielding 1,000 generated videos per viewpoint and 6,000 additional videos per subset. Each generated video inherits the split assignment, gloss annotation, and sentence annotation of its source clip, and sampling is performed independently across viewpoints while keeping all generated variants within the split of their source clips. 
MVSign-H2S is derived from the naturally captured multi-view videos in How2Sign~\citep{duarte2021how2sign}. We retain all original frontal-view videos and native $\mathrm{R90}^{\circ}$ videos, and sample 100 clips from each original train, dev, and test split to synthesize each of the five additional viewpoints $\{\mathrm{R45}^{\circ},\mathrm{L30}^{\circ},\mathrm{L60}^{\circ},\mathrm{U30}^{\circ},\mathrm{D30}^{\circ}\}$. This produces 300 generated videos per viewpoint, with 100 videos assigned to each split, and each video inherits the split assignment and sentence annotation of its source clip. Since How2Sign does not provide gloss annotations, MVSign-H2S is evaluated only for gloss-free SLT, whereas MVSign-PT14 and MVSign-CSL are evaluated for CSLR, gloss-based SLT, and gloss-free SLT.

\subsection{Evaluation Metric.}
Following prior SLU works~\citep{chen2022simple,tang2024signidd,gan2024signgraph,tang2025gloss,wang2025linguistics}, we adopt the Word Error Rate (WER) metric for the CSLR task, and ROUGE (ROUGE-L) and BLEU-1-4 for SLT tasks. Hereafter, ROUGE refers to ROUGE-L.

\textbf{WER.} WER measures the minimum number of word-level substitutions, deletions, and insertions required to transform the predicted gloss sequence into the reference sequence:
\begin{equation}
\mathrm{WER}
=
\frac{S+D+I}{N},
\end{equation}
where $S$, $D$, and $I$ denote the numbers of substitutions, deletions, and insertions, respectively, and $N$ is the number of tokens in the reference gloss sequence. Lower WER indicates better recognition performance.

\textbf{BLEU.} BLEU-$n$ measures the modified precision of matched $n$-grams between
a generated sentence $\hat{S}$ and a reference sentence $S$. The modified
$n$-gram precision is defined as
\begin{equation}
p_n
=
\frac{
\sum_{g}
\min\left\{
\mathrm{Count}_{\hat{S}}(g),
\mathrm{Count}_{S}(g)
\right\}
}{
\sum_{g}\mathrm{Count}_{\hat{S}}(g)
}.
\end{equation}
BLEU-4 is computed as
\begin{equation}
\mathrm{BLEU}\text{-}4
=
\mathrm{BP}
\exp\left(
\frac{1}{4}\sum_{n=1}^{4}\log p_n
\right),
\end{equation}
where $\mathrm{BP}$ denotes the brevity penalty:
\begin{equation}
\mathrm{BP}
=
\begin{cases}
1, & c>r,\\
\exp\left(1-\frac{r}{c}\right), & c\leq r,
\end{cases}
\end{equation}
with $c$ and $r$ denoting the lengths of the generated and reference
sentences, respectively. BLEU-1, BLEU-2, and BLEU-3 are computed analogously
using $n=1$, $n=2$, and $n=3$. Higher BLEU scores indicate better translation
quality.

\textbf{ROUGE-L.} ROUGE-L is based on the length of the longest common subsequence (LCS) between
the generated sentence $\hat{S}$ and the reference sentence $S$. Let
$L=\mathrm{LCS}(\hat{S},S)$, $c=|\hat{S}|$, and $r=|S|$. The LCS-based precision
and recall are defined as
\begin{equation}
P_{\mathrm{LCS}}=\frac{L}{c},
\qquad
R_{\mathrm{LCS}}=\frac{L}{r}.
\end{equation}
ROUGE-L is then computed as their weighted harmonic mean:
\begin{equation}
\mathrm{ROUGE\text{-}L}
=
\frac{
(1+\gamma^2)P_{\mathrm{LCS}}R_{\mathrm{LCS}}
}{
R_{\mathrm{LCS}}+\gamma^2P_{\mathrm{LCS}}
},
\end{equation}
where $\gamma$ controls the relative importance of recall and precision.
Higher ROUGE-L scores indicate better translation quality.

\subsection{Implementation Details.}
\label{sub_Implementation_Details}
We adopt the same data preprocessing as in previous work~\citep{gan2024signgraph,gan2025mixsigngraph} for CanonSLU. Specifically, we use ResNet-18 as the visual encoder. All input videos are resized to $224\times224$ pixels and sampled with a frame interval of one. The models are optimized using Adam with an initial learning rate of $1\times10^{-4}$ and a weight decay of $1\times10^{-4}$. For SLT, the learning rates of the translation module and the visual embedding layer are set to $1\times10^{-5}$ and $1\times10^{-3}$, respectively. The batch size is set to 2, and CanonSLU is trained for 50 epochs.
We use a MultiStep learning-rate schedule with milestones at epochs 20 and 35, reducing the learning rate by a factor of 0.2 at each milestone. For CKT, the distillation temperature and loss weight are set to $T_d=8$ and $\lambda=40$, respectively. 
MRA is inserted after the third and fourth stages of the visual encoder. Its temporal offsets are set to $\delta\in\{1,2,4\}$, with $k=24$ and $k=12$ selected spatial positions at the third and fourth stages, respectively.
All experiments are implemented using PyTorch Distributed Data Parallel on 8 NVIDIA RTX 4080S GPUs.

\section{Additional Experimental Results}
\label{sec:supp_Additional Experimental Results}

\begin{table*}[h]
\centering
\caption{Ablation results of different module combinations for Gloss-based SLT on MVSign-PT14.}
\label{tab:Ablation results of different module combinations for Gloss-based SLT on MVSign-PT14.}
\renewcommand{\arraystretch}{1.3}
\resizebox{\textwidth}{!}{
\begin{tabular}{ccc ccccc ccccc}
\toprule
 \multicolumn{3}{c}{\textbf{Methods}} &\multicolumn{5}{c}{\textbf{DEV}} &\multicolumn{5}{c}{\textbf{TEST}} \\
\cmidrule(l){4-8}
\cmidrule(l){9-13}
\textbf{Baseline} &\textbf{CKT} &\textbf{MRA} &\makecell{\textbf{BLEU1$\uparrow$}} &\makecell{\textbf{BLEU2$\uparrow$}} &\makecell{\textbf{BLEU3$\uparrow$}} &\makecell{\textbf{BLEU4$\uparrow$}} &\makecell{\textbf{ROUGE$\uparrow$}} &\makecell{\textbf{BLEU1$\uparrow$}} &\makecell{\textbf{BLEU2$\uparrow$}} &\makecell{\textbf{BLEU3$\uparrow$}} &\makecell{\textbf{BLEU4$\uparrow$}} &\makecell{\textbf{ROUGE$\uparrow$}} \\
\midrule
\ding{51} &&
&41.76 &29.69 &21.94 &17.98 &\multicolumn{1}{c|}{41.67}
&40.23 &26.29 &19.92 &15.52 &39.85 \\
\ding{51} &\ding{51} &
&43.28 &30.16 &22.90 &18.18 &\multicolumn{1}{c|}{43.61}
&42.47 &29.24 &21.91 &17.19 &41.66 \\
\ding{51} & &\ding{51}
&44.69 &31.94 &24.56 &19.73 &\multicolumn{1}{c|}{45.12}
&43.56 &30.88 &23.78 &18.86 &44.76 \\
\cellcolor{lightgray}\ding{51} &\cellcolor{lightgray}\ding{51} &\cellcolor{lightgray}\ding{51}
&\cellcolor{lightgray}\textbf{46.66} &\cellcolor{lightgray}\textbf{33.68} &\cellcolor{lightgray}\textbf{26.11} &\cellcolor{lightgray}\textbf{21.19} &\multicolumn{1}{c|}{\cellcolor{lightgray}\textbf{46.89}}
&\cellcolor{lightgray}\textbf{44.54} &\cellcolor{lightgray}\textbf{31.98} &\cellcolor{lightgray}\textbf{24.53} &\cellcolor{lightgray}\textbf{19.73} &\cellcolor{lightgray}\textbf{45.47} \\
\bottomrule
\end{tabular}
}
\end{table*}

\textbf{Effectiveness of CKT and MRA for Gloss-based SLT.}
Table~\ref{tab:Ablation results of different module combinations for Gloss-based SLT on MVSign-PT14.} evaluates the contributions of CKT and MRA to gloss-based SLT on MVSign-PT14.
Both components improve translation performance individually, increasing test BLEU-4 by 1.67 and 3.34 points over the baseline, respectively.
Combining them achieves the best results across all reported metrics, reaching test BLEU-4 and ROUGE scores of 19.73 and 45.47, with gains of 4.21 and 5.62 points.
These results suggest that canonical-view guidance during CSLR pretraining and motion-aware feature learning provide complementary benefits for downstream translation.

\begin{table*}[h]
\renewcommand\arraystretch{1.2}
\caption{Comparison of viewpoint-wise CSLR performance (WER$\downarrow$) on the paired-source MVSign-PT14 test set. 
}
\label{tab:Comparison of viewpoint-wise CSLR performance on the paired-source MVSign-PT14 test set.}
\centering
\resizebox{0.7\textwidth}{!}{
\begin{tabular}{c cccccccc}
\toprule
   \textbf{Methods} &\textbf{All} &\textbf{R90$^\circ$} &\textbf{L60$^\circ$} &\textbf{R45$^\circ$} &\textbf{L30$^\circ$} &\textbf{D30$^\circ$} &\textbf{U30$^\circ$} &\textbf{Front}\\
\midrule
    SMKD~\citep{hao2021self} &53.55 &63.52 &54.72 &52.20 &55.37 &55.97 &54.72 &38.36\\
    VAC~\citep{min2021visual} &51.12 &62.87 &55.36 &50.94 &52.83 &54.09 &50.94 &30.82\\
    SEN~\citep{hu2023self} &50.40 &58.49 &53.46 &50.91 &50.94 &54.08 &53.46 &31.45\\
    CorrNet~\citep{hu2023continuous}  &46.06 &55.09 &47.91 &47.82 &48.23 &45.13 &47.03 &31.19\\
    SignGraph~\citep{gan2024signgraph} &45.73 &55.77 &46.40 &47.25 &48.48 &44.14 &47.13 &30.92\\
\midrule
    \cellcolor{lightgray}CanonSLU (Ours) &\cellcolor{lightgray}\textbf{44.65} &\cellcolor{lightgray}\textbf{53.46} &\cellcolor{lightgray}\textbf{45.71} &\cellcolor{lightgray}\textbf{45.28} &\cellcolor{lightgray}\textbf{47.84} &\cellcolor{lightgray}\textbf{43.35} &\cellcolor{lightgray}\textbf{46.17} &\cellcolor{lightgray}\textbf{30.79} \\
\bottomrule
\end{tabular}}
\end{table*}

\textbf{Paired-source viewpoint evaluation.}
Table~\ref{tab:Comparison of viewpoint-wise CSLR performance on the paired-source MVSign-PT14 test set.} reports CSLR performance on the paired-source MVSign-PT14 test set.
We randomly select 100 frontal-view clips from the original test split and generate all six non-frontal views for each clip, preserving their split assignments and annotations. 
CanonSLU achieves the lowest WER on the overall test set and every viewpoint. Its overall WER is 44.65\%, improving over the strongest baseline, SignGraph (45.73\%), by 1.08 points. The largest gains are observed at R45$^\circ$ and R90$^\circ$, with reductions of 1.97 and 1.63 points over the strongest baseline for each view, respectively. CanonSLU also improves performance on the pitch views, reducing WER by 0.79 points at D30$^\circ$ and 0.86 points at U30$^\circ$. The consistent improvements across views, including the frontal view, show that the method remains effective when source-clip content is controlled across viewpoints.

\begin{table*}[h]
\renewcommand\arraystretch{1.2}
\caption{Comparison of viewpoint-wise CSLR performance (WER$\downarrow$) on MVSign. 
}
\label{tab:Comparison of viewpoint-wise CSLR performance on MVSign.}
\centering
\resizebox{\textwidth}{!}{
\begin{tabular}{c cccccccc cccccccc}
\toprule
   \multirow{2}{*}{\textbf{Methods}} &\multicolumn{8}{c}{\textbf{MVSign-PT14 (Test set)}} &\multicolumn{8}{c}{\textbf{MVSign-CSL (Test set)}}\\
   \cmidrule(l){2-9}\cmidrule(l){10-17}
   &\textbf{All} &\textbf{R90$^\circ$} &\textbf{L60$^\circ$} &\textbf{R45$^\circ$} &\textbf{L30$^\circ$} &\textbf{D30$^\circ$} &\textbf{U30$^\circ$} &\textbf{Front} &\textbf{All} &\textbf{R90$^\circ$} &\textbf{L60$^\circ$} &\textbf{R45$^\circ$} &\textbf{L30$^\circ$} &\textbf{D30$^\circ$} &\textbf{U30$^\circ$} &\textbf{Front} \\
\midrule
    VAC~\citep{min2021visual} &39.57 &61.47 &49.63 &54.82 &56.29 &51.95 &49.38 &26.36   &44.59 &64.78 &61.21 &58.70 &58.54 &64.43 &57.54 &36.37\\
    SEN~\citep{hu2023self} &38.17 &61.93 &49.63 &53.29 &51.57 &49.40 &50.16 &24.81     &41.11 &62.01 &60.81 &55.78 &55.36 &59.82 &52.30 &32.75\\
    CorrNet~\citep{hu2023continuous}  &34.80 &56.42 &46.41 &49.00 &49.38 &44.14 &47.81 &22.61   &40.21 &64.38 &60.67 &53.52 &52.19 &62.98 &52.69 &31.36\\
    SignGraph~\citep{gan2024signgraph} &35.13 &57.92 &48.13 &49.62 &49.33 &48.74 &48.10 &22.95  &37.63 &57.96 &56.69 &53.20 &51.27 &58.60 &48.21 &29.95\\
\midrule
    \cellcolor{lightgray}CanonSLU (Ours) &\cellcolor{lightgray}\textbf{33.35} &\cellcolor{lightgray}\textbf{52.37} &\cellcolor{lightgray}\textbf{45.31} &\cellcolor{lightgray}\textbf{47.01} &\cellcolor{lightgray}\textbf{47.95} &\cellcolor{lightgray}\textbf{43.74} &\cellcolor{lightgray}\textbf{46.66} &\cellcolor{lightgray}\textbf{20.66} &\cellcolor{lightgray}\textbf{35.71} &\cellcolor{lightgray}\textbf{55.67} &\cellcolor{lightgray}\textbf{54.77} &\cellcolor{lightgray}\textbf{51.00} &\cellcolor{lightgray}\textbf{49.93} &\cellcolor{lightgray}\textbf{55.60} &\cellcolor{lightgray}\textbf{47.70} &\cellcolor{lightgray}\textbf{27.26}\\
\bottomrule
\end{tabular}}
\end{table*}

\textbf{View-wise CSLR Performance.}
Table~\ref{tab:Comparison of viewpoint-wise CSLR performance on MVSign.} reports the test-set WER for each viewpoint on MVSign-PT14 and MVSign-CSL. CanonSLU achieves the lowest WER in every reported column, including the aggregate score, all six non-frontal views, and the frontal view. On MVSign-PT14, CanonSLU reduces the aggregate WER to 33.35\%, improving over the best listed baseline, CorrNet (34.80\%), by 1.45 points. The largest view-wise gain is observed at R90$^\circ$, where the WER decreases from 56.42\% to 52.37\%. On MVSign-CSL, CanonSLU achieves an aggregate WER of 35.71\%, which is 1.92 points lower than SignGraph (37.63\%), and consistently improves performance across all viewpoints. The gains are particularly evident for D30$^\circ$ (3.00 points), R90$^\circ$ (2.29 points), and the frontal view (2.69 points). These results indicate that CanonSLU improves recognition across both horizontal and pitch variations rather than relying on a single viewpoint.

\begin{table*}[h]
\renewcommand\arraystretch{1.1}
\caption{Effect of temporal offset sets in MRA on MVSign-PT14.}
\label{tab:Effect of temporal offset sets in MRA on MVSign-PT14.}
\centering
\resizebox{0.7\textwidth}{!}{
\begin{tabular}{c ccc ccc}
\toprule
   \multirow{2}{*}{\textbf{$\mathcal{T}_{\mathrm{off}}$}} &\multicolumn{3}{c}{\textbf{DEV}} &\multicolumn{3}{c}{\textbf{TEST}}\\
   \cmidrule(l){2-4}\cmidrule(l){5-7}
   &\textbf{WER$\downarrow$} &\textbf{BLEU-4$\uparrow$} &\textbf{ROUGE$\uparrow$} &\textbf{WER$\downarrow$} &\textbf{BLEU-4$\uparrow$} &\textbf{ROUGE$\uparrow$}\\
\midrule
    $\{1\}$ &34.09 &20.13 &45.38 &34.55 &18.62 &44.09\\
    $\{1,2\}$ &34.99 &19.96 &45.54 &35.35 &18.37 &44.12\\
    \cellcolor{lightgray}$\{1,2,4\}$ &\cellcolor{lightgray}\textbf{33.27} &\cellcolor{lightgray}\textbf{21.19} &\cellcolor{lightgray}\textbf{46.89} &\cellcolor{lightgray}\textbf{33.35} &\cellcolor{lightgray}\textbf{19.73} &\cellcolor{lightgray}\textbf{45.47}\\
    $\{1,2,4,8\}$ &34.54 &20.17 &45.06 &34.64 &18.03 &44.37\\
\bottomrule
\end{tabular}}
\end{table*}

\textbf{Selection of Temporal Offsets.}
Table~\ref{tab:Effect of temporal offset sets in MRA on MVSign-PT14.} compares different temporal offset sets for MRA.
The configuration $\mathcal{T}_{\mathrm{off}}=\{1,2,4\}$ achieves the lowest WER and highest BLEU-4 and ROUGE on both splits, supporting its effectiveness for CSLR and gloss-based SLT.
Performance does not improve monotonically as more offsets are included.
Extending the set to $\{1,2,4,8\}$ increases test WER from 33.35 to 34.64 and reduces test BLEU-4 from 19.73 to 18.03.
These results support the default offset set $\{1,2,4\}$ for temporal relation aggregation.

\begin{table*}[h]
\renewcommand\arraystretch{1.1}
\caption{Effect of the top-k values of MRA at stages 3 and 4 on MVSign-PT14.}
\label{tab:Effect of the top-k values of MRA at stages 3 and 4 on MVSign-PT14.}
\centering
\resizebox{0.7\textwidth}{!}{
\begin{tabular}{cc ccc ccc}
\toprule
   \multirow{2}{*}{\textbf{stage3}} &\multirow{2}{*}{\textbf{stage4}} &\multicolumn{3}{c}{\textbf{DEV}} &\multicolumn{3}{c}{\textbf{TEST}}\\
   \cmidrule(l){3-5}\cmidrule(l){6-8}
   & &\textbf{WER$\downarrow$} &\textbf{BLEU-4$\uparrow$} &\textbf{ROUGE$\uparrow$} &\textbf{WER$\downarrow$} &\textbf{BLEU-4$\uparrow$} &\textbf{ROUGE$\uparrow$}\\
\midrule
    24 &6 &34.80 &19.98 &45.10 &34.95 &19.37 &44.25\\
    \cellcolor{lightgray}24 &\cellcolor{lightgray}12 &\cellcolor{lightgray}\textbf{33.27} &\cellcolor{lightgray}\textbf{21.19} &\cellcolor{lightgray}\textbf{46.89} &\cellcolor{lightgray}\textbf{33.35} &\cellcolor{lightgray}\textbf{19.73} &\cellcolor{lightgray}\textbf{45.47}\\
    24 &24 &34.71 &20.20 &44.72 &34.77 &19.44 &44.06\\
\midrule
    12 &12 &34.75 &20.08 &44.39 &34.77 &18.84 &44.01\\
    \cellcolor{lightgray}24 &\cellcolor{lightgray}12 &\cellcolor{lightgray}\textbf{33.27} &\cellcolor{lightgray}\textbf{21.19} &\cellcolor{lightgray}\textbf{46.89} &\cellcolor{lightgray}\textbf{33.35} &\cellcolor{lightgray}\textbf{19.73} &\cellcolor{lightgray}\textbf{45.47}\\
    36 &12 &35.16 &19.67 &43.80 &35.33 &18.66 &43.26\\
\bottomrule
\end{tabular}}
\end{table*}

\textbf{Effect of Top-$k$ Selection.}
Table~\ref{tab:Effect of the top-k values of MRA at stages 3 and 4 on MVSign-PT14.} examines the number of spatial positions selected at each MRA stage.
We vary the top-$k$ value at one stage while keeping the other fixed.
Setting $k=24$ at stage 3 and $k=12$ at stage 4 achieves the best dev results across all three metrics, together with the lowest test WER and highest test ROUGE.
The configuration $(12,12)$ yields a slightly lower test BLEU-4 of 18.84, compared with 19.73 for $(24,12)$.
Overall, these results support $(24,12)$ as the default configuration based on its validation performance and strong results across tasks.

\begin{table}[t]
\centering
\caption{Perceptual evaluation of synthesized videos.
Scores range from 1 to 5, with higher values indicating better quality.
Each source-generated pair is rated by five participants.
The acceptance rate denotes the percentage of pairs whose mean rating across participants is at least 4.0 for each of the three criteria.}
\label{tab:quality_Evaluation}
\resizebox{\linewidth}{!}{
\begin{tabular}{lcccccc}
\toprule
\textbf{Subset} &\textbf{Views} &\textbf{Pairs} &\textbf{Visual realism} &\textbf{Temporal coherence}
&\textbf{Motion consistency} &\textbf{Acceptance} \\
\midrule
MVSign-PT14 &6 synthetic views &60 (10/view) &4.18 &4.06 &4.02 &86.67\% \\
MVSign-CSL  &6 synthetic views &60 (10/view) &4.11 &4.00 &3.95 &85.00\% \\
MVSign-H2S  &5 synthetic views &50  (10/view) &4.29 &4.18 &4.15 &86.00\% \\
\midrule
Overall & -- &170 &4.19 &4.07 &4.03 &85.88\% \\
\bottomrule
\end{tabular}}
\end{table}

\textbf{Perceptual Evaluation.}
Table~\ref{tab:quality_Evaluation} reports the perceptual quality of synthesized videos, evaluated by five participants on 170 source-generated pairs, with 10 pairs sampled per synthetic viewpoint in each subset.
The synthesized videos achieve overall scores of 4.19 for visual realism, 4.07 for temporal coherence, and 4.03 for motion consistency on a five-point scale.
Moreover, 85.88\% of the pairs receive average ratings of at least 4.0 on all three criteria, with acceptance rates ranging from 85.00\% to 86.67\% across the subsets.
MVSign-H2S obtains the highest mean scores across all three criteria, while motion consistency receives the lowest overall rating, particularly on MVSign-CSL.
These results suggest that the generated videos generally exhibit favorable \textbf{Visual realism}, \textbf{Temporal coherence}, and \textbf{Motion consistency} across viewpoints.

\begin{CJK}{UTF8}{gbsn} 
\begin{table}[h]
\renewcommand\arraystretch{1.3}
\caption{CSLR qualitative results on MVSign-PT14 and MVSign-CSL. We use different colors to represent \textcolor{green}{substitutions}, \textcolor{blue}{deletions}, and \textcolor{red}{insertions}, respectively.}
\label{tab:CSLR qualitative results}
\centering
\resizebox{\textwidth}{!}{
\begin{tabular}{l|c}
\toprule
    \textbf{Example (a)} &\textbf{MVSign-PT14}\\
\midrule
    Groundtruth &IX KOENNEN GEFAHR GLATT BESONDERS MITTE REGION SUED DANN KALT\\
    VAC &IX KOENNEN \textcolor{blue}{GEFAHR} GLATT BESONDERS MITTE REGION SUED \textcolor{green}{SONNE} KALT\\
    SignGraph &\textcolor{red}{\_\_ON\_\_} IX KOENNEN GEFAHR \textcolor{blue}{GLATT} BESONDERS MITTE REGION SUED DANN KALT\\
    CanonSLU (Ours) &IX KOENNEN GEFAHR GLATT BESONDERS MITTE REGION SUED DANN KALT\\
\toprule
    \textbf{Example (b)} &\textbf{MVSign-PT14}\\
\midrule
    Groundtruth &OST REGION SONNE MEHR WARM MAXIMAL SECHS ZWANZIG SUED MAXIMAL SIEBEN ZWANZIG GRAD\\
    VAC &OST REGION SONNE MEHR WARM \textcolor{blue}{MAXIMAL} SECHS ZWANZIG SUED MAXIMAL SIEBEN ZWANZIG GRAD\\
    SignGraph &OST REGION SONNE MEHR WARM MAXIMAL SECHS \textcolor{green}{KOENNEN} SUED MAXIMAL \textcolor{blue}{SIEBEN} ZWANZIG GRAD\\
    CanonSLU (Ours) &OST REGION SONNE MEHR WARM MAXIMAL SECHS ZWANZIG SUED MAXIMAL SIEBEN ZWANZIG GRAD\\
\toprule
    \textbf{Example (c)} &\textbf{MVSign-CSL}\\
\midrule
    Groundtruth &这 看 手机 这 女 他 你 女 朋友 是\\
    VAC &这 看 手机 这 女 \textcolor{blue}{他} 你 女 朋友 是\\
    SignGraph &这 \textcolor{blue}{看} 手机 这 女 他 \textcolor{red}{多少} 你 女 朋友 是\\
    CanonSLU (Ours) &这 看 手机 这 女 他 你 女 朋友 是\\
\toprule
    \textbf{Example (d)} &\textbf{MVSign-CSL}\\
\midrule
    Groundtruth &小 张 这 女 是 你 公司 你 了解 他\\
    VAC &小 张 这 女 是 \textcolor{green}{他} 公司 你 了解 他\\
    SignGraph &小 张 这 女 是 你 \textcolor{blue}{公司} 你 了解 他\\
    CanonSLU (Ours) &小 张 这 女 是 你 公司 你 了解 他\\
\bottomrule
\end{tabular}}
\end{table}
\end{CJK}

\textbf{Qualitative Results on CSLR.}
Table~\ref{tab:CSLR qualitative results} compares recognition outputs on MVSign-PT14 and MVSign-CSL.
VAC and SignGraph exhibit gloss substitutions, deletions, and insertions.
For example, VAC omits ``GEFAHR'' and substitutes ``DANN'' with ``SONNE'' in example (a), while SignGraph replaces ``ZWANZIG'' with ``KOENNEN'' and misses ``SIEBEN'' in example (b).
The MVSign-CSL examples similarly reveal pronoun confusion and omitted glosses.
CanonSLU correctly recovers all four reference sequences, illustrating more accurate gloss recognition and better sequence completeness in these examples.

\begin{CJK}{UTF8}{gbsn} 
\begin{table}[h]
\renewcommand\arraystretch{1.3}
\caption{SLT qualitative results on MVSign.}
\label{tab:SLT qualitative results}
\centering
\resizebox{\textwidth}{!}{
\begin{tabular}{l|c}
\toprule
    \textbf{Example (a)} &\textbf{MVSign-PT14}\\
\midrule
    Groundtruth &im verlauf kommt im norden und nordwesten teilweise kräftiger regen auf einzelne gewitter sind dabei\\
    MMTLB &im verlauf kommt im norden und teilweise kräftiger regnet auf einzelne gewitter sind dabei\\
    CV-SLT &im verlauf kommt im norden wolken nordwesten teilweise kräftiger sonntag auf einzelne gewitter sind dabei\\
    CanonSLU (Ours) &im verlauf kommt im norden und nordwesten teilweise kräftiger regen auf einzelne gewitter sind dabei\\
\toprule
    \textbf{Example (b)} &\textbf{MVSign-CSL}\\
\midrule
    Groundtruth &我 给 你 预 订 了 一 个 生 日 蛋 糕 。\\
    MMTLB &我 给 你 订 了 一 个 生 日 蛋 糕 在 超 市。\\
    CV-SLT &我 给 你 买 一 个 生 日 蛋 糕 。\\
    CanonSLU (Ours) &我 给 你 预 订 了 一 个 生 日 蛋 糕 。\\
\toprule
    \textbf{Example (c)} &\textbf{MVSign-H2S}\\
\midrule
    Groundtruth &We're going to slide the front foot and drag the back foot whenever you're moving forward\\
    GFSLT-VLP &We're going to slide the front foot and drag what good foot whenever you're moving forward\\
    SpaMo &We're going to slide the stance foot and drag the back foot whenever you're came forward\\
    CanonSLU (Ours) &We're going to slide the front foot and drag the back foot whenever you're moving forward\\
\bottomrule
\end{tabular}}
\end{table}
\end{CJK}

\textbf{Qualitative Results on SLT.}
Table~\ref{tab:SLT qualitative results} presents translation examples from the three MVSign subsets.
In example (a), MMTLB omits ``nordwesten'', while CV-SLT replaces ``regen'' with ``sonntag'', altering the weather description.
In example (b), MMTLB adds an unsupported supermarket location, whereas CV-SLT changes the action from reserving a cake to buying one.
In example (c), GFSLT-VLP mistranslates the back-foot reference, while SpaMo changes the front-foot reference and produces an ungrammatical description of forward movement.
CanonSLU matches the reference translations in all three examples, preserving the geographic information, action meanings, and spatial relations.


\end{document}